\documentclass[review]{elsarticle}

\usepackage{lineno,hyperref}
\modulolinenumbers[5]
\usepackage{color}
\usepackage{diagbox}
\usepackage{subfigure}
\usepackage{multirow}
\usepackage{enumerate}
\usepackage{rotating}
\usepackage{graphicx}
\usepackage{algorithm}
\usepackage{algpseudocode}
\usepackage{amsfonts}
\usepackage{amsmath}
\usepackage{amssymb}
\usepackage{epsfig}
\usepackage{epstopdf}
\usepackage{mathrsfs}
\usepackage[misc]{ifsym}

\graphicspath{{./figures/}}
\newcommand{\etal}{\textit{et al.}}
\journal{Journal of Pattern Recognition}

\begin{document}

\begin{frontmatter}
\title{Deep Fisher Discriminant Learning for Mobile Hand Gesture Recognition}
%\title{Elsevier \LaTeX\ template\tnoteref{mytitlenote}}
%\tnotetext[mytitlenote]{Fully documented templates are available in the elsarticle package on \href{http://www.ctan.org/tex-archive/macros/latex/contrib/elsarticle}{CTAN}.}

%% Group authors per affiliation:
\author[mymainaddress]{Chunyu Xie}%\fnref{myfootnote}}
%\address{Department of Automation, Beihang University, Beijing, China}
%\fntext[myfootnote]{Since 1880.}

\author[mysecondaryaddress]{Ce Li}
%\ead[url]{www.elsevier.com}

\author[mymainaddress]{Baochang Zhang\corref{mycorrespondingauthor}}
\cortext[mycorrespondingauthor]{Corresponding author}
\ead{bczhang@buaa.edu.cn}

\author[mythirdaddress]{Chen Chen}
\author[myfourthaddress]{Jungong Han}

\address[mymainaddress]{Department of Automation, Beihang University, Beijing, China}
\address[mysecondaryaddress]{Department of Computer Science, China University of Mining and Technology, Beijing, China}
\address[mythirdaddress]{University of Central Florida, Orlando, FL, USA.}
\address[myfourthaddress]{Nortumbria Univesity, Newcastle, UK.}

%---------------------------------------------------------
%---------------------------------------------------------
\begin{abstract}
Gesture recognition is a challenging problem in the field of biometrics. In this paper, we integrate Fisher criterion into Bidirectional Long-Short Term Memory (BLSTM) network and Bidirectional Gated Recurrent Unit (BGRU), thus leading to two new deep models termed as F-BLSTM and F-BGRU. Both Fisher discriminative deep models can effectively classify the gesture based on analyzing the acceleration and angular velocity data of the human gestures. Moreover, we collect a large Mobile Gesture Database (MGD) based on the accelerations and angular velocities containing 5547 sequences of 12 gestures. Extensive experiments are conducted to validate the superior performance of the proposed networks as compared to the state-of-the-art BLSTM and BGRU on MGD database and two benchmark databases (\emph{i.e.} BUAA mobile gesture and SmartWatch gesture).
\end{abstract}


\end{frontmatter}

%\linenumbers
%---------------------------------------------------------
\section{Introduction}\label{sec:introduction}
Towards natural human-computer interaction, the emergence of smartphones has changed our lives and made our lives more convenient.  The interaction between people and mobile phones are through the touch screen, camera and microphone. However, due to environmental constraints, these interaction methods suffer from many uncontrollable problems. For example, the video based methods do not work well in the night time. %Thus, other forms of interaction such as voice and gestures have gained widespread attention.
Nowadays, inertial sensors including accelerometer and gyrometer are built in smartphones \cite{lane2010a,choi2005,mantyla2000hand}, which can record the signal of hand movements when their devices are in use. Therefore, gesture action using inertial sensors can be easily achieved and understood \cite{liu2009uwave}. %For gesture input, the key component is the correct recognition of gestures. Gesture recognition is essential to deal with human gestures and accept the purpose of people.
Different from computer vision based gesture recognition, the requirements for gesture recognition based on inertial sensors (\emph{e.g.} accelerometer and gyrometer) are much simpler\;\cite{catal2015on}. However, the mobile gesture recognition inevitably encounters several external variations including signal intensity differences (intense versus weak gestures), temporal variations (slow versus fast movements) and physical differences (users' physical conditions, etc.). In addition, the noise caused by the hardware device also has a severe impact on the recognition performance. Support vector machine (SVM), hidden Markov model (HMM) and dynamic time warping (DTW) have been introduced to solve the problems mentioned above.

Recently, deep learning methods have significantly push the state-of-the-art in human activity recognition. They avoid feature engineering and are able to learn data representations and classifiers simultaneously. Among those, the recurrent neural network (RNN) and Long Short-Term Memory (LSTM)\;\cite{hochreiter1997long} are popular sequential modelling methods. They have been successfully applied to many fields such as language modeling\;\cite{mikolov2011extensions,sundermeyer2012lstm,mesnil2013investigation}, image captioning\;\cite{vinyals2015show,xu2015icml}, video analysis\;\cite{ng2015cvpr,alahi2016social,deng2015structure,ibrahim2015a} and 3D action recognition\;\cite{,du2015hierarchical,veeriah2015differential,wang2014learning,liu2016eccv}. These methods can also be utilized to solve our problem, since the mobile gesture signals are sequential data streams of inertial sensors\;\cite{shin2016dynamic,lefebvre2013ANN}.

The publicly available databases are very important to researchers for algorithm development and evaluation. However, most of the previous works in the field of mobile gesture recognition were conducted with self-prepared data. And few databases related to mobile gesture recognition are publicly available. In this paper, we first introduce a mobile based gesture recognition benchmark, which can help researchers to evaluate and compare their algorithms by using the same gesture data. The database was collected based on 32 participants (23 males and 9 females), consisting 12 gestures of 5547 repetitions.

\begin{figure*}[t]
	\normalsize
	\centering
	\includegraphics[width=\linewidth,height=0.4\textheight]{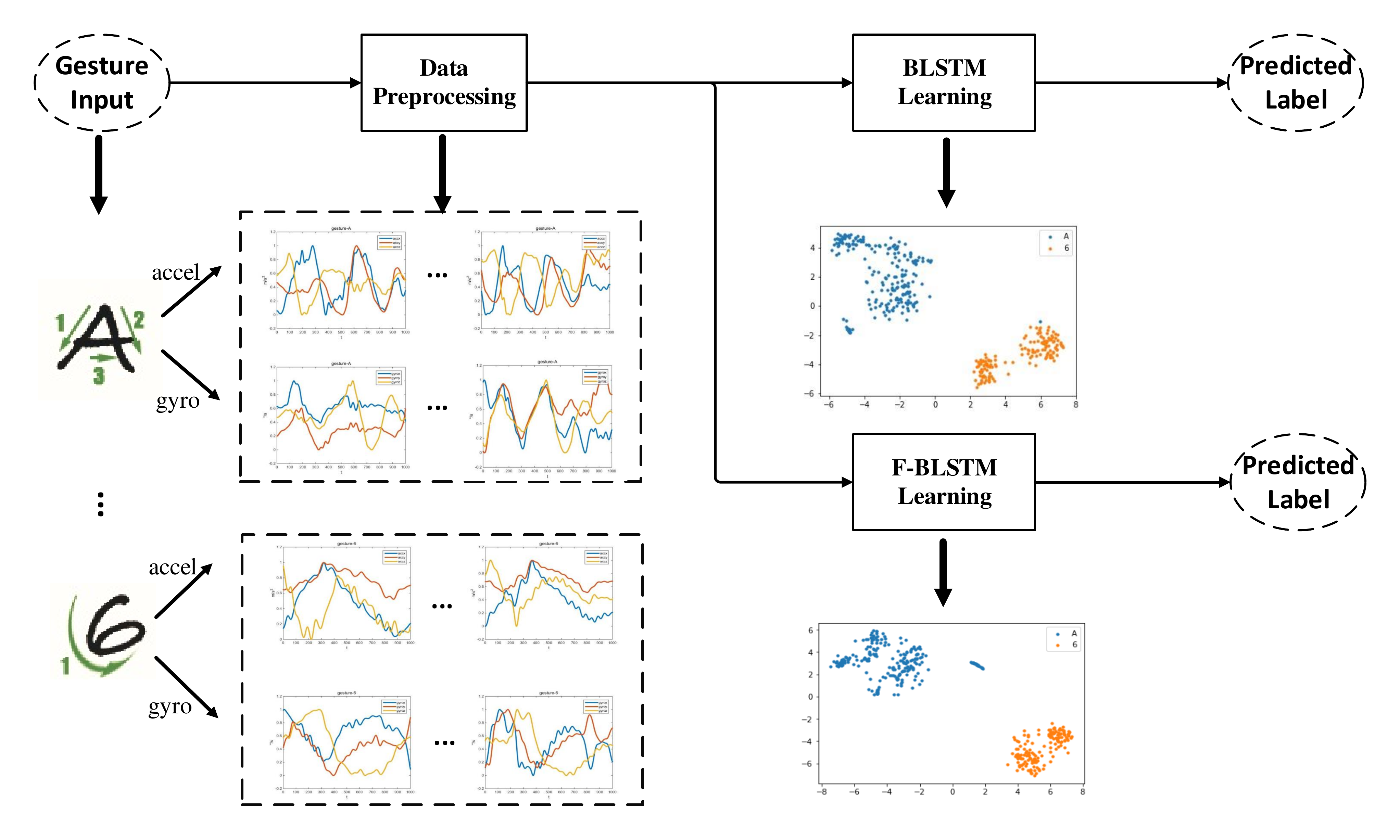}
	\caption{Flowchart of the gesture recognition system. We introduce Fisher criterion into BLSTM network to improve the traditional softmax loss training function, which is able to minimize the intra-class variations and maximize the inter-class variations in the deep framework. }
	\label{fig:flowchart}%fig.1
	%\vspace*{4pt}
\end{figure*}

Two Fisher discriminative deep models termed F-BLSTM and a variant F-BGRU are proposed for hand gesture recognition on mobile devices.
The Fisher criterion which minimizes the {intra-class variations and maximize inter-class variations} is incorporated into softmax loss of LSTM, leading a better classification ability to cope with external variations. The flowchart of the gesture recognition approach is shown in Fig.\;\ref{fig:flowchart}.

In summary, we make the following contributions:

\begin{enumerate}[1.]
	\item We collect a large mobile gesture database using an Andriod {Huawei} device, which is the largest database in published studies for mobile gesture recognition systems.
	\item We integrate Fisher criterion into BLSTM network to improve the traditional softmax loss training function. Extensive experiments on our {MGD, BUAA Mobile Gesture database,} and a public database are conducted to verify the superior performance of the proposed networks.
\end{enumerate}

The rest of the paper is organized as follows. Section\;\ref{sec:relatedwork} introduces the related works, and Section\;\ref{sec:approach} describes the details of the proposed method. Experiments and results are presented in Section\;\ref{sec:experiments}. Finally, Section\;\ref{sec:conclusion} concludes the paper.

%---------------------------------------------------------
\section{Related Work}\label{sec:relatedwork}
 {Gesture recognition has been extensively investigated in the last two decades, and remarkable advances are achieved using inertial sensors in mobile devices\;\cite{rekimoto2001iswc,jang2003signal,kallio2003online,bulling2014a}. For the application of human computer interaction, Rekimoto \etal\;\cite{rekimoto2001iswc} detected the movement of arm using a specific wearable device. But it is difficult to get high precision in practice because of large size of the equipment. Fresca \etal \;\cite{ferscha2007gestural} studied and recognized human gesticulation and the manipulation of graspable and movable everyday artifacts as a potentially effective means for the interaction with smart things. Parsani \etal \;\cite{Parsani2009A} designed an embedded system which could analyze and recognize smartphone gestures involving a combination of straight line motions in three dimensions. Roy \etal \;\cite{Roy2014Demo} suggested that the walking direction should be detectable through the accelerometer and get blended into various other motion patterns during the act of walking, including up and down bounce, side-to-side sway, swing of arms or legs, \emph{etc.}. They also analyzed the human walking dynamics to estimate the dominating forces and used this knowledge to find the heading direction of the pedestrian. For the application of biological monitoring, Park \etal\;\cite{Park2014Poster} demonstrated a very promising application to classify and monitor heartbeats, while Nandakumar \etal \;\cite{Nandakumar2015Contactless} monitored sleep apnea using the sensors in smartphones to develop more convenient conditions for gesture recognition\;\cite{hoang2013adaptive}. Besides, there are also some other applications.} As reported in\;\cite{Agrawal2011Using}, Agrawal \etal  proposed a system called PhonePoint Pen that uses the built-in accelerometer in mobile phones to recognize the human writing. The system, based on Nokia N95 platform, was evaluated through 10 students and 5 hospital patients. Results showed that English characters can be identified with an average accuracy of 91\%. The system presented a promising prospect for mobile based gesture recognition.
 %Cho \etal\;\cite{cho2005two-stage} collected acceleration signals by Samsung cell phones. Researchers use various methods to achieve good classification results.

{Mobile gesture recognition has provided new directions and also delivered compelling performance for the application of machine learning. Hofmann \etal\;\cite{hofmann1998velocity} proposed a recognition scheme based on Hidden Markov Models (HMM)\;\cite{markov} and used discrete HMM (dHMM) to recognize dynamic gestures. The approach essentially divided the input data into different regions and assigned each of them to a corresponding codebook for classifying them with dHMM. The vector codebook was obtained by a clustering method, serving as an unsupervised learning procedure to model the feature vector distribution in the input data space. The experiments were carried out using 500 training gestures with 10 samples per gesture, yielding an accuracy of 95.6\% for 100 test gestures.} In\;\cite{kallio2003online}, gestures were captured with a small wireless sensor-box that produced three dimensional acceleration signal. Kallio \etal trained the dHMM model by using five states and a codebook size of eight. They {measured} the recognition accuracy of system using four degrees of complexity. In\;\cite{kela2006accelerometer-based}, an HMM model was trained with five states, achieving a rate of 96.1\% accuracy for classifying 8 gestures. Pylv \etal\;\cite{pylvanainen2005accelerometer} proposed a method based on continuous HMM (cHMM), which takes correlated time information into consideration. The experiment achieved reliable results, with 96.67\% of correct classification on a database of 20 samples for 10 gestures. In\;\cite{Zhang2009Hand}, multi-stream HMM consisting of EMG and ACC streams was utilized as decision fusion method to recognize hand gestures. For a data set of 18 gestures, each trained with 10 repetitions, the average recognition accuracy was about 91.7\% in real application.

{Besides HMM, a few other popular techniques have been used in gesture recognition. Akl \etal\; proposed a gesture recognition system based primarily on a single 3 dimensional accelerometer, by employing DTW\;\cite{akl2010accelerometer-based}.} The system defined a dictionary of 18 gestures and a database of 3700 repetitions from 7 users{ and got up to 90\% classification accuracy in the experiment.} David \etal\;\cite{Mace2013Accelerometer} proposed two approaches including Naive Bayes and DTW for recognizing four gesture types from five different subjects in the experiment. The results revealed Bayesian classification is better than DTW. Wu \etal\;\cite{wu2009gesture} employed multi-class Support Vector Machine (SVM) for user-independent gesture recognition and demonstrated that SVM significantly outperformed other methods including DTW, Naive Bayes and HMM. In\;\cite{Hsu2009Integrating}, Wang \etal~combined LCS and SVM to perform the classification task and achieved the classification accuracy of 93\%. Another line of research focuses on the feature extraction and selection. For example, the principle component analysis was used for feature selection and dimensionality reduction in gesture classification\;\cite{Marasovic2011Accelerometer}. {In\;\cite{He2011Accelerometer}, the hybrid features combined short-time energy with Fast Fourier Transform, denoting the fusion of time-domain features and frequency-domain features, were presented for recognizing seventeen complex gestures on cell phone.}
An average recognition accuracy of 89.89\% was obtained using multi-class SVM.

{Driven by the tremendous success of deep learning, the research paradigm has been shifted from traditional approaches to deep learning methods for mobile gesture recognition\;\cite{shin2016dynamic,lefebvre2013ANN}.} Shin \etal\;\cite{shin2016dynamic} developed a dynamic hand gesture recognition technique using recurrent neural network (RNN) algorithm. The gesture recognition model was trained using the SmartWatch Gestures database\;\cite{costante2014eusipco}. Each gesture sequence contains acceleration data from the 3 dimensional accelerometer. An evaluation of the network size was presented, and the best performance was obtained by using the LSTM layer with the size of 128.  Lefebvre\;\cite{lefebvre2013ANN} carried out gesture recognition experiments on a database consisted of both accelerometer and gyrometer sensors. The sensor data was captured using an Android Nexus S Samsung device. 22 participants, from 20 to 55 years old, contributed to the database of the 14 symbolic gestures. The results showed that gesture recognition utilizing both sensors can achieve better performance than using each individual sensor. {Moreover, the BLSTM based method achieved an accuracy of 95.57\% on the database of total 1540 gestures. To the best of our knowledge, the BLSTM based method is currently the state-of-the-art baseline and performs better than previous approaches such as cHMM, DTW, FDSVM and LSTM}.

%---------------------------------------------------------
\section{The Proposed Approach}\label{sec:approach}
In this section, we first describe the network structures of {Bidirectional Long-Short Term Memory (BLSTM) and its variant -- LSTM with Gate Recurrent Unit (BGRU).} Then, we propose to incorporate the Fisher criterion to improve the discriminative power of these deep models, dubbed F-BLSTM and F-BGRU.

%---------------------------------------------------------
\subsection{Bidirectional LSTM}\label{sec:blstm}
\begin{figure*}
	\centering
	%\begin{minipage}[t]{0.6\linewidth} height=0.9\textheight
	\includegraphics[height=0.9\textheight]{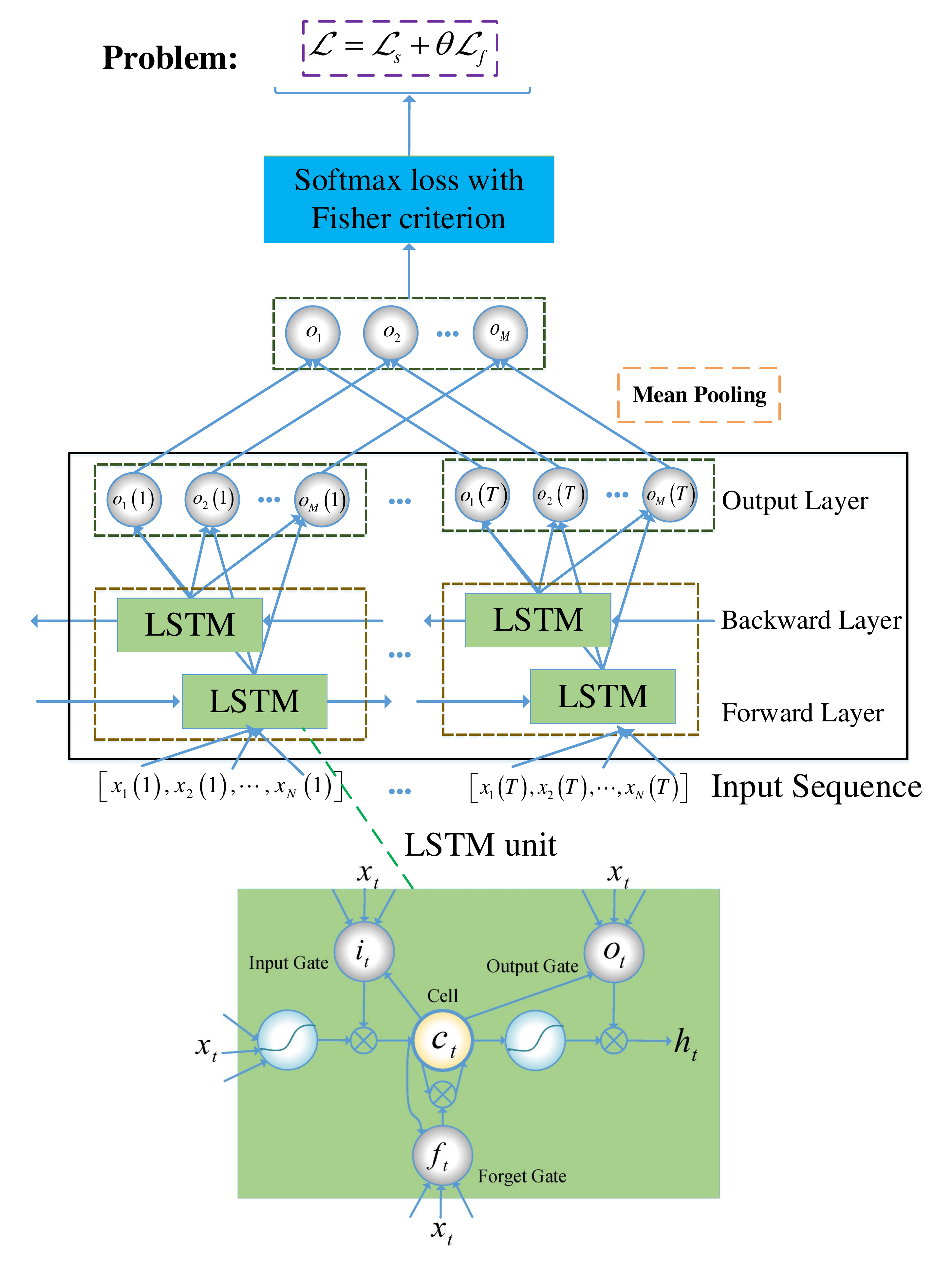}
	%[width=\linewidth]
	\caption{{The architecture of F-BLSTM. We intuitively change the loss function of BLSTM, and the resulting algorithm does not affect the training convergence and the model size, but leading to a performance improvement.}}
	\label{fig:blstm-architecture}
	%\end{minipage}	
\end{figure*}

We briefly describe the LSTM unit which is the basic building block of the proposed F-BLSTM model. The neurons of LSTM contain a constant memory cell name, which has a state $c_t$ at time $t$. {A LSTM neuron unit is presented in detail at the bottom of Fig.\;\ref{fig:blstm-architecture}}. Each LSTM unit for reading or modifying is controlled through a sigmoidal input gate $i_t$, a forget gate $f_t$ and an output gate $o_t$. At each time step $t$, {LSTM unit receives inputs from two external sources at each of the three gates.} The current frame $x_t$ and previous hidden states $h_{t-1}$ are two sources, and the cell state $c_{t-1}$ in the cell block is an internal source of each gate. The gates are passed through the tanh non-linearity and activated by logistic function. After multiplying the cell state by the forget gate $f_t$, the final output of the LSTM unit $h_t$ is computed by multiplying the activation $o_t$ of the output gates with updated cell state. Denoting all $W_*$ are diagonal matrices, the updates in a layer of LSTM units are summarized as follows:
\begin{equation}
	\begin{array}{l}
	{i_t} = \sigma \left( {{W_{xi}}{x_t} + {W_{hi}}{h_{t - 1}} + {W_{ci}}{c_{t - 1}} + {b_i}} \right),\\
	{f_t} = \sigma \left( {{W_{xf}}{x_t} + {W_{hf}}{h_{t - 1}} + {W_{cf}}{c_{t - 1}} + {b_f}} \right),\\
	{c_t}={f_t}c_{t-1}+{i_t}tanh\left(W_{xc}x_t+W_{hc}h_{t-1}+b_c\right),\\
	o_t=\sigma\left(W_{xo}x_t+W_{ho}h_{t-1}+W_{co}c_t+b_o\right),\\
	h_t={o_t}tanh\left(c_t\right).
	\end{array}
	\label{eq:lstm-unit}
\end{equation}

The model of {BLSTM uses Recurrent Neural Nets (RNNs)} made of LSTM units, which have shown the great ability to deal with temporal data in many applications~\cite{sundermeyer2012lstm}. We consider the gesture data using 3 dimensional accelerometer and 3 dimensional gyrometer synchronized input vectors through sampling timestep. As shown in Fig.\;\ref{fig:blstm-architecture}, the forward and backward LSTM hidden layers are fully connected to the input layer and consist of multiple LSTM neurons each with full recurrent connections. Several experiments have been conducted with different hidden neuron sizes and 128 neurons yield the best results. The output layer has a size equivalent to the number of neuron to classify (\emph{i.e.} $M=128$). $G = \left\{ {G_1,...,G_{T}} \right\}$ is a gesture sequence of $T$ size, $G_t = \left(x_1\left(t\right),...,x_{N}\left(t\right)\right)$ is a vector at timestep $t$ and $N$ denotes the sensor number, and $\left(y_1,...,y_{n}\right)$ is the BLSTM output set with $n$ being the number of gestures to be classified. The softmax activation function is used for this layer to give network response between 0 and 1. Classically, these outputs can be considered as posterior probabilities of the input sequence belonging to a specific gesture class, and the softmax loss function is presented as follows
\begin{equation}
	{\cal L}_s =  - \frac{1}{m} \sum\limits_{i = 1}^m \log \frac{{{e^{W_{{y_i}}^T{O_i} + {b_{{y_{i}}}}}}}}{{\sum\nolimits_{j = 1}^n {{e^{W_j^T{O_i} + {b_j}}}} }},
	\label{eq:softmaxloss}
\end{equation}
where $O_i =\left(o_1,...,o_M\right)$ denotes the {$i$th output belonging to the $y_i$th class}. $W_j$ denotes the $j$th column of the weights $W$ in the last layer and $b$ is the bias term. $m$ is the size of mini-batch and $n$ is the number of classes.

%\begin{figure*}[htbp]
%	\centering
%	\includegraphics[width=0.8\linewidth]{blstm-framework.jpg}
%	\caption{The framework of BLSTM.}
%	\label{fig:blstm-framework}
%\end{figure*}

%---------------------------------------------------------
\subsection{Deep Fisher Discriminant Learning }\label{sec:fisherloss}

To further enhance the performance of BLSTM, we incorporate the Fisher criterion into the softmax loss function, which is shown in Fig.\;\ref{fig:blstm-architecture}. First, the input layer consists of the concatenation of 3 dimensional accelerometer and 3 dimensional gyrometer signals synchronized in time (\emph{i.e.} $N=6$). The sensor data is normalized between 0 and 1 according to the maximum value that sensors can provide.
In order to minimize the intra-class variations and maximize the inter-class variations of gesture data, we propose a new Fisher criterion based on Fisher Linear Discrimination as follows:
\begin{equation}
	{\cal L}_f
	= \frac{1}{m}{\sum\limits_{i = 1}^m {\left\| {{O_i} - {\mu_{{y_i}}}} \right\|_2^2}  - \frac{\delta}{n\left(n-1\right)} \sum\limits_{j = 1,k = 1}^n {\left\| {{\mu_j} - {\mu_k}} \right\|_2^2} }
	\label{eq:fisher}
\end{equation}
where ${\mu_{{y_i}}}$ is the $y_i$th class mean of output vectors, and $\delta$ is the discriminative factor. As updating the mean vector ${\mu_{{y_i}}}$ when learning BLSTM, the Fisher criterion utilizes the whole training set and mean vectors of each class in every iteration. We propose to augment the loss in Eq.\;(\ref{eq:softmaxloss}) with the additional Fisher criterion term in Eq.\;(\ref{eq:fisher}) as follows:
\begin{equation}
	%\begin{aligned}
	{\cal L} 
	= {\cal L}_s +\theta{\cal L}_f \\
%	%= &\frac{1}{m}{\sum\limits_{i = 1}^m {\left\| {{O_i} - {\mu_{{y_i}}}}  \right\|_2^2}  - \frac{\delta}{n\left(n-1\right)} \sum\limits_{j = 1,k = 1}^n {\left\| {{\mu_j} - {\mu_k}} \right\|_2^2} 
%	%\end{aligned}	
%	\begin{array}{l}
%	{\cal L} = {\cal L}_s +\theta{\cal L}_f \\
%	\;\;\; = \frac{1}{m}\sum\limits_{i = 1}^m {\left\| {{O_i} - {\mu _{{y_i}}}} \right\|_2^2}  - \frac{\delta }{{n\left( {n - 1} \right)}}\sum\limits_{j = 1,k = 1}^n {\left\| {{\mu _j} - {\mu _k}} \right\|_2^2} 
%	\end{array}
	\label{eq:wholeloss}
\end{equation}
where $\theta$ and $\delta$ are bounded in $[0,1]$, and these two parameters are used for balancing three parts of the loss function. In forward and backward processes, we set output vector ${O_i}$, mean vector ${\mu_j}$, loss parameter $W$, scalar parameters $\theta$, $\delta$ and learning rate $\lambda$, BLSTM parameters $H_f$ and iteration number $e$, respectively. In each iteration, we compute the loss of F-BLSTM by Eq.~(\ref{eq:fisher}) and Eq.~(\ref{eq:wholeloss}), and the backpropagation error by
\begin{equation}
	\frac{{\partial {L^e}}}{{\partial O_i^e}} = \frac{{\partial L_s^e}}{{\partial O_i^e}} + \theta \frac{{\partial L_f^e}}{{\partial O_i^e}}.
	\label{eq:backerror}
\end{equation}
Then, we update the parameter $W$, mean vector $\mu_{j}$ and BLSTM parameter $H_f$ in the $e+1$ iteration by the following formulas until the converge stopping criterion.
\begin{equation}
	\begin{array}{l}
	{W^{e + 1}} = {W^e} - {\lambda ^e} \cdot \frac{{\partial L_f^e}}{{\partial {W^e}}},\\
	\mu _j^{e + 1} = \mu _j^e - \alpha  \cdot \Delta \mu _j^e,\\
	H_f^{e + 1} = H_f^e - {\lambda ^e}\sum\nolimits_i^m {\frac{{\partial {L^e}}}{{\partial O_i^e}}}  \cdot \frac{{\partial O_i^e}}{{\partial H_f^e}}.
	\end{array}
	\label{eq:parmean}
\end{equation}
%Here, the learning details of new F-BLSTM (Fisher discriminative BLSTM) model are summarized in Algorithm\;\ref{alg:fd-blstm}.

With proper scalar parameters $\theta$, $\delta$ and $\alpha$, the discriminative power of F-LSTM can be significantly enhanced for hand gesture recognition. This network is learned using classical online backpropagation through time with momentum. For classifying a new gesture sequence, we use a majority voting rule over the outputs along the sequence (\emph{i.e.} keeping only the most probable class $argmax _{i \in \left[ {1,n} \right]}{O_i}$) to determine the final gesture class. A detailed parameter analysis of $\theta$, $\delta$ and $\alpha$ is presented in Section\;\ref{sec:parameter}.

%\begin{algorithm}[h]
%	\caption{F-BLSTM learning algorithm.}
%	\label{alg:fd-blstm}
%	\begin{algorithmic}[1]
%		\State Set vector ${O_i}$, mean vector ${\mu_j}$, loss parameter $W$, scalar paramters $\theta$, $\delta$, and $\alpha$, learning rate $\lambda$, and BLSTM parameters $H_f$, respectively.
%		\State Initialize number of iteration $e=1$.
%		\Repeat
%		\For {$e= 1 : maximum$}
%		\State compute the loss of F-BLSTM by ${\cal L} ={\cal L}_s +\theta{\cal L}_f$,
%		\State ${\cal L}={\cal L}_s + \theta \sum\limits_{i = 1}^m {\left\| {{O_i} - {\mu_{{y_i}}}} \right\|_2^2}  - \theta \cdot \delta \sum\limits_{j = 1,k = 1}^n {\left\| {{\mu_j} - {\mu_k}} \right\|_2^2}$.
%		\State compute the backpropagation error $\frac{{\partial {L^e}}}{{\partial O_i^e}} = \frac{{\partial L_s^e}}{{\partial O_i^e}} + \theta \frac{{\partial L_f^e}}{{\partial O_i^e}}$
%		\State update the loss parameter $W$: ${W^{e + 1}} = {W^e} - {\lambda ^e} \cdot \frac{{\partial {\cal L}_f^e}}{{\partial {W^e}}}$,
%		\State 	mean vector $\mu_{j}$: $\mu_{j}^{e+1} =\mu_{j}^{e}-\alpha \cdot \Delta {\mu_{j}^{e}}$,
%		\State 	BLSTM parameter $H_f$: $H_f^{e+1}   =H_f^{e}- {\lambda ^e}\sum\nolimits_i^m {\frac{{\partial {L^e}}}{{\partial O_i^e}}}  \cdot \frac{{\partial O_i^e}}{{\partial H_f^e}} $.
%		\EndFor
%		\State $e = e+1$
%		\Until{converge stopping criterion}
%	\end{algorithmic}
%\end{algorithm}
%---------------------------------------------------------
\subsection{Bidirectional GRU and F-BGRU}\label{sec:bgru}
\begin{figure*}[h]
	\centering
	%\begin{minipage}[t]{0.6\linewidth}
	\includegraphics[width=0.65\linewidth]{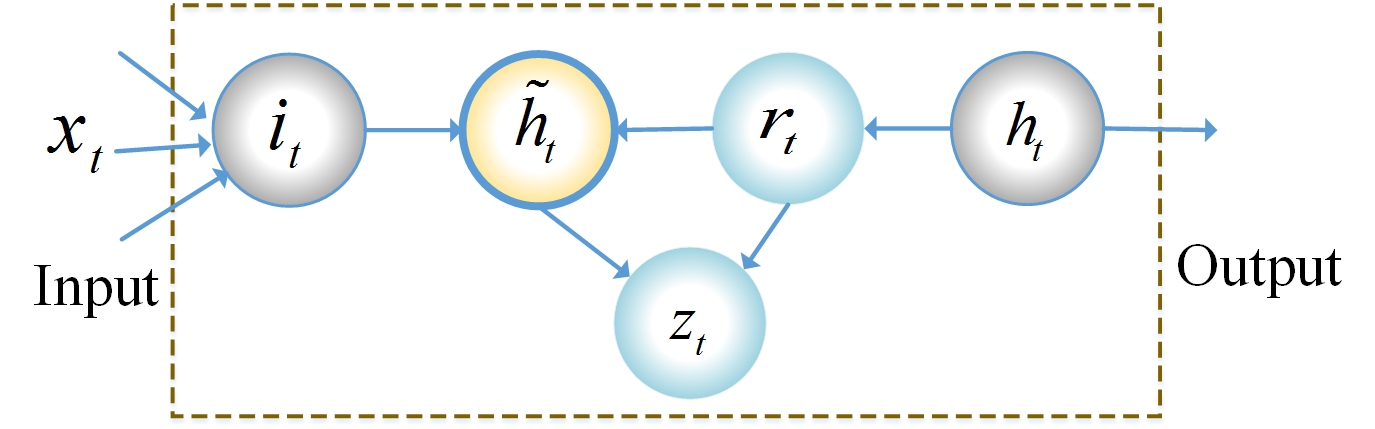}
	\caption{BGRU unit.}
	\label{fig:bgru-unit}
	%\end{minipage}
\end{figure*}

To further enhance the performance of network, a variant of BLSTM termed Bidirectional Gated Recurrent Unit (BGRU) was proposed in\;\cite{cho2014arxiv,chung2014eprint} to make each recurrent unit to adaptively capture dependence of different time scales. Similarly to the BLSTM unit, the BGRU has the activation $h_t$, candidate activation $\tilde{h}_t$, update gate $z_t$ and reset gate $r_t$ units to modulate the flow of information in unit without some separate memory cells, as shown in Fig.\;\ref{fig:bgru-unit}. The flows in BGRU are summarized as follows:

\begin{equation}
\begin{array}{l}
{z_t} = \sigma \left( {{W_z}{x_t} + {W_{zf}}{h_t} + {b_z}} \right),\\
{r_t} = \sigma \left( {{W_r}{x_t} + {W_{rf}}{h_t} + {b_r}} \right),\\
{{\tilde h}_t} = tanh\left( {W{x_t} + U\left( {{r_t} \odot {h_{t - 1}}} \right) + {b_h}} \right),\\
{h_t} = \left( {1 - {z_t}} \right){h_{t - 1}} + {z_t}{{\tilde h}_t}
\end{array}.
\label{eq:bgru-unit}
\end{equation}
where the activation $h_t$ at time $t$ is a linear interpolation between previous activation $h_{t-1}$ and the candidate activation $\tilde{h}_t$, the candidate activation $\tilde{h}_t$ is computed same as traditional recurrent unit. The update gate $z_t$ decides the number of units to update its activation, and so as the reset gate $r_t$.

It is easy to notice that the BGRU unit also controls the flow of information like the BLSTM unit, but without having to use a memory unit. Similar to F-BLSTM, we also apply the Fisher discriminative function for BGRU and learn a new variant named F-BGRU model to recognize hand gestures.

%---------------------------------------------------------
\section{Experiments}\label{sec:experiments}
%---------------------------------------------------------
\subsection{Hardware Device}\label{sec:database}
Our mobile hand gesture database is collected using a Huawei device with Android system, which has a 3 dimensional accelerometer and a 3 dimensional gyrometer. According to\;\cite{lefebvre2013ANN}, we collect the data of both accelerometer and gyrometer, {and record each gesture by pressing, holding and releasing the ``Sensor" button on the touch screen.}

%\textbf{Gesture Dictionary.}
%\paragraph{Gesture Dictionary}
\subsubsection{Gesture Dictionary}
\begin{figure*}[htbp]
	\normalsize
	\centering
	\includegraphics[width=\linewidth]{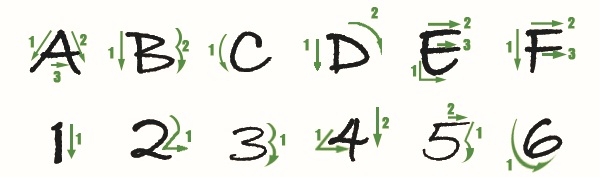}
	\caption{Examples of hand gestures in \textbf{{MGD} database}.}
	\label{fig:gesture-dic}%fig.4
	\vspace*{4pt}
\end{figure*}

As shown in Fig.\;\ref{fig:gesture-dic}, the gesture dictionary consists of two categories including Arabic numerals (1, 2, 3, 4, 5, 6) and English capital letters (A, B, C, D, E, F). Furthermore, the stroke order of gestures is set in advance to ensure the consistency of gestures drawed by left or right hand of each participant.
%However, this will result in some unnecessary continuous strokes, which are inevitable when drawing characters. For example, when a participant performs a ``D" operation, one has to return the phone to the starting point in the second step after completing the first step.
We directly collect all accelerometer and gyrometer data, then transfer the data from the cellphone memory to the computer for gesture recognition.

%\textbf{Database Collection.}
%\paragraph{Database Collection}
\subsubsection{Database Collection} The database named MGD consists of 12 gestures performed by 32 participants (23 males and 9 females) with about fifteen times per gesture. Therefore, there are a total of 5547 gesture sequences. The sampling time of accelerometer and gyrometer sensors is 5ms corresponding to a frequency of 200Hz. To the best of our knowledge, it is the largest database so far for mobile based gesture recognition, {which is of benefit to the research community}.

%---------------------------------------------------------
\subsection{Implementation Details}\label{sec:preprocess}%Data Preprocessing
In this section, we present the details of the implementation of our experiments. We use Tensorflow toolbox as the deep learning platform and an NVIDIA GTX 1070 GPU to run the experiments. In order to validate the effectiveness of our proposed Fisher criterion in LSTM for modeling temporal sequences, we compare our methods, F-BLSTM and F-BGRU, with the state-of-the-art baselines (BLSTM and BGRU\;\cite{chung2014eprint}) on three benchmarks including our proposed database (MGD), and two previous databases: the BUAA Mobile Gesture database\;\cite{xie2016ccbr} and the SmartWatch Gestures database\;\cite{chung2014eprint}. We comprehensively evaluate the performance of the proposed model under different parameter settings of $\delta$, $\alpha$ and $\theta$ in Sec.\;\ref{sec:parameter}, and provide extensive experimental comparison in Sec.\;\ref{sec:comparison}.

\begin{figure*}[htbp]
	\normalsize
	\centering
	\subfigure[Original Accelerometer Data]{
		\centering
		\includegraphics[width=0.48\linewidth]{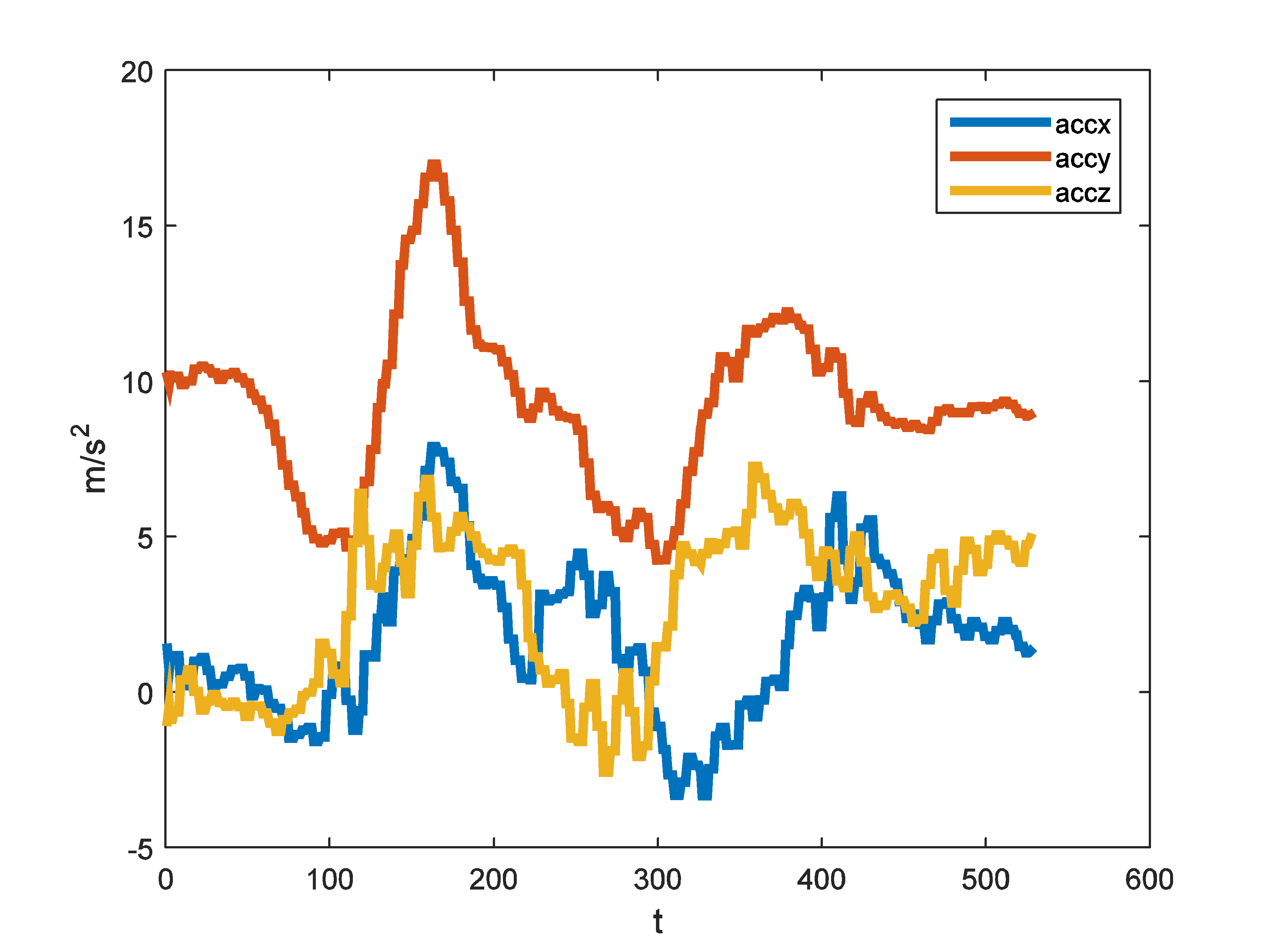}}
	\subfigure[Original Gyrometer Data]{
		\centering
		\includegraphics[width=0.48\linewidth]{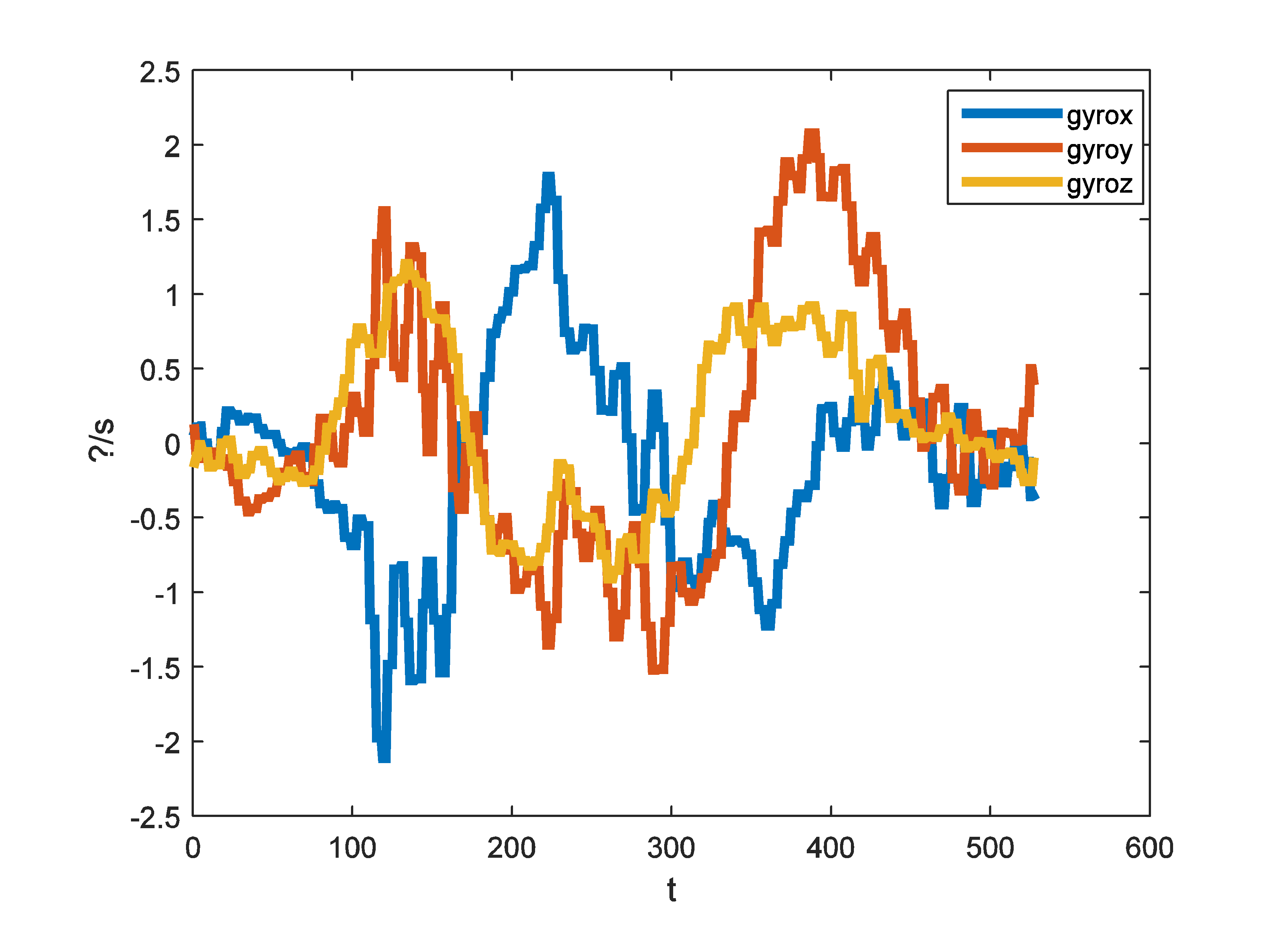}}
	\\
	\subfigure[Filtered Accelerometer Data]{
		\centering
		\includegraphics[width=0.48\linewidth]{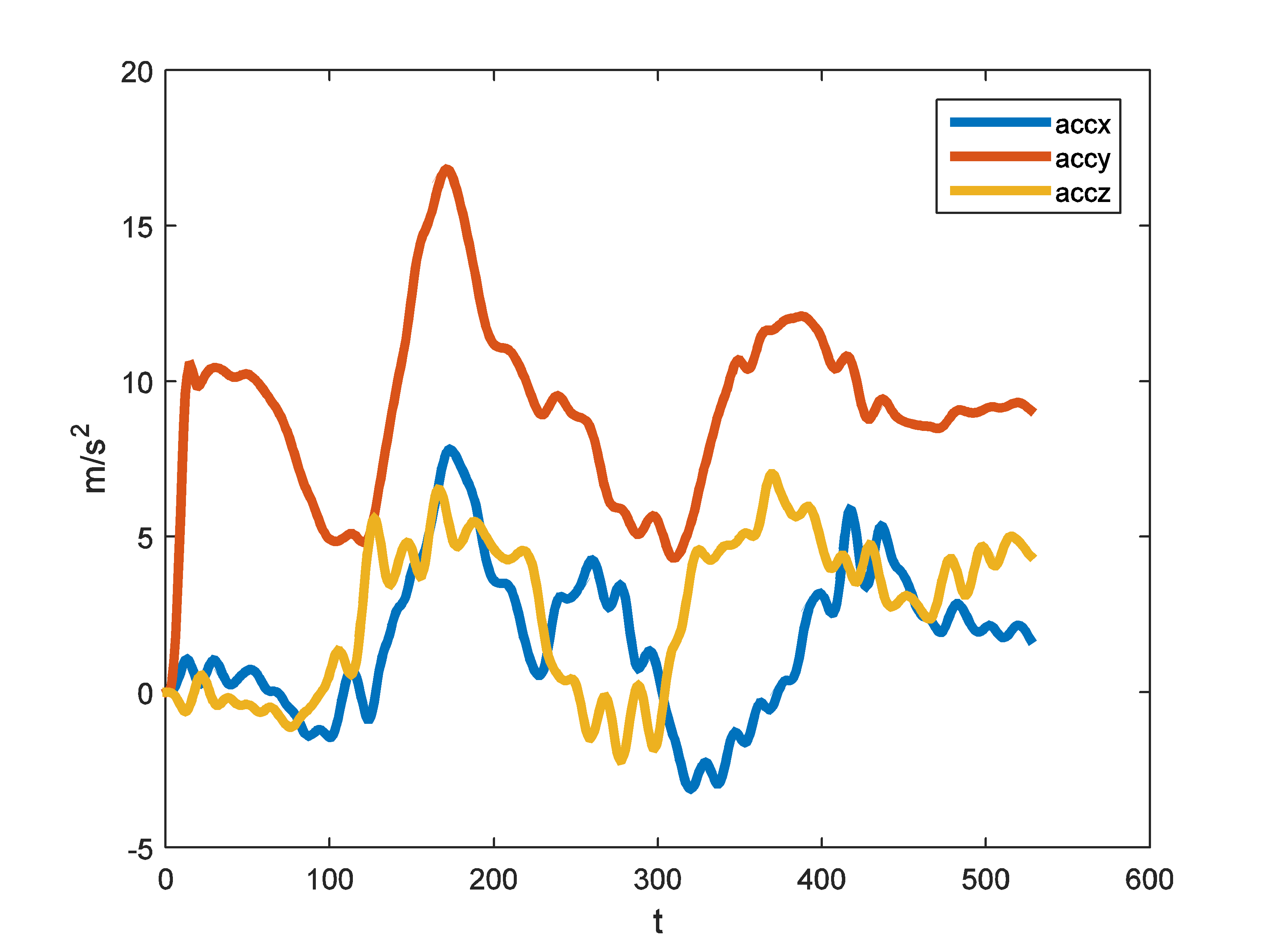}}
	\subfigure[Filtered Gyrometer Data]{
		\centering
		\includegraphics[width=0.48\linewidth]{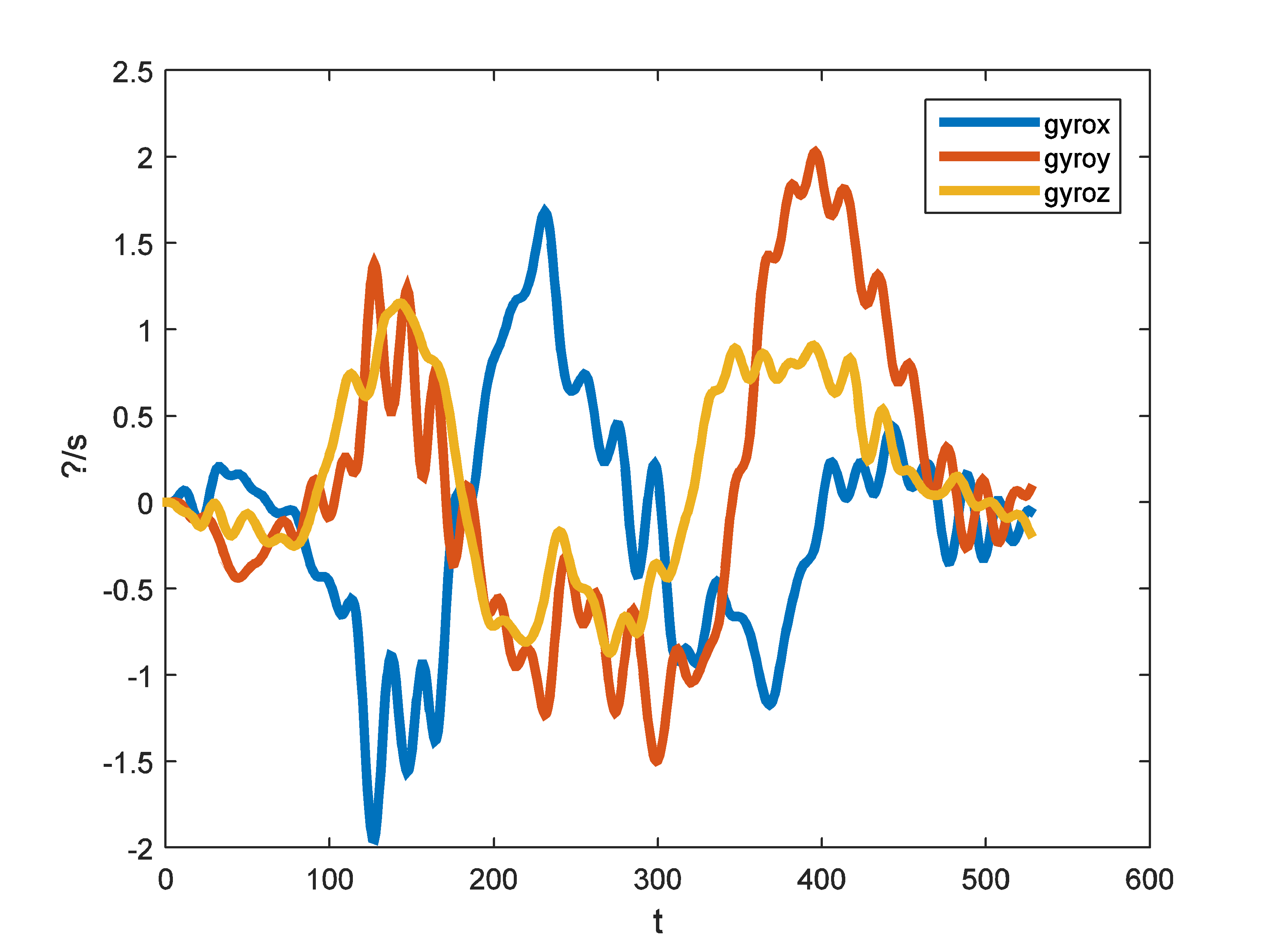}}
	\\
	\caption{The original accelerometer and gyrometer data vs. the processed data by Moving Average Filter.}
	\label{fig:filtered-data}
	\vspace*{4pt}
\end{figure*}

\textbf{Data preprocessing.} The main objective for data preprocessing is to facilitate gesture recognition. In real world applications, the sensor data often contain a lot of noise due to complex environmental conditions and hardware limitations. Therefore, we first carry out a filtering process to suppress noise (i.e. data smoothing). We experimented with Average Filter, Median Filter and Butterworth Filter, and selected the Average Filter in terms of its good performance and computational efficiency. Fig.\;\ref{fig:filtered-data} shows the original accelerometer and gyrometer data and the processed data using the Average Filter.

\begin{figure*}[htbp]
	\normalsize
	\centering
	\subfigure[Preprocessed Accelerometer Data]{
		\centering
		\includegraphics[width=0.47\linewidth]{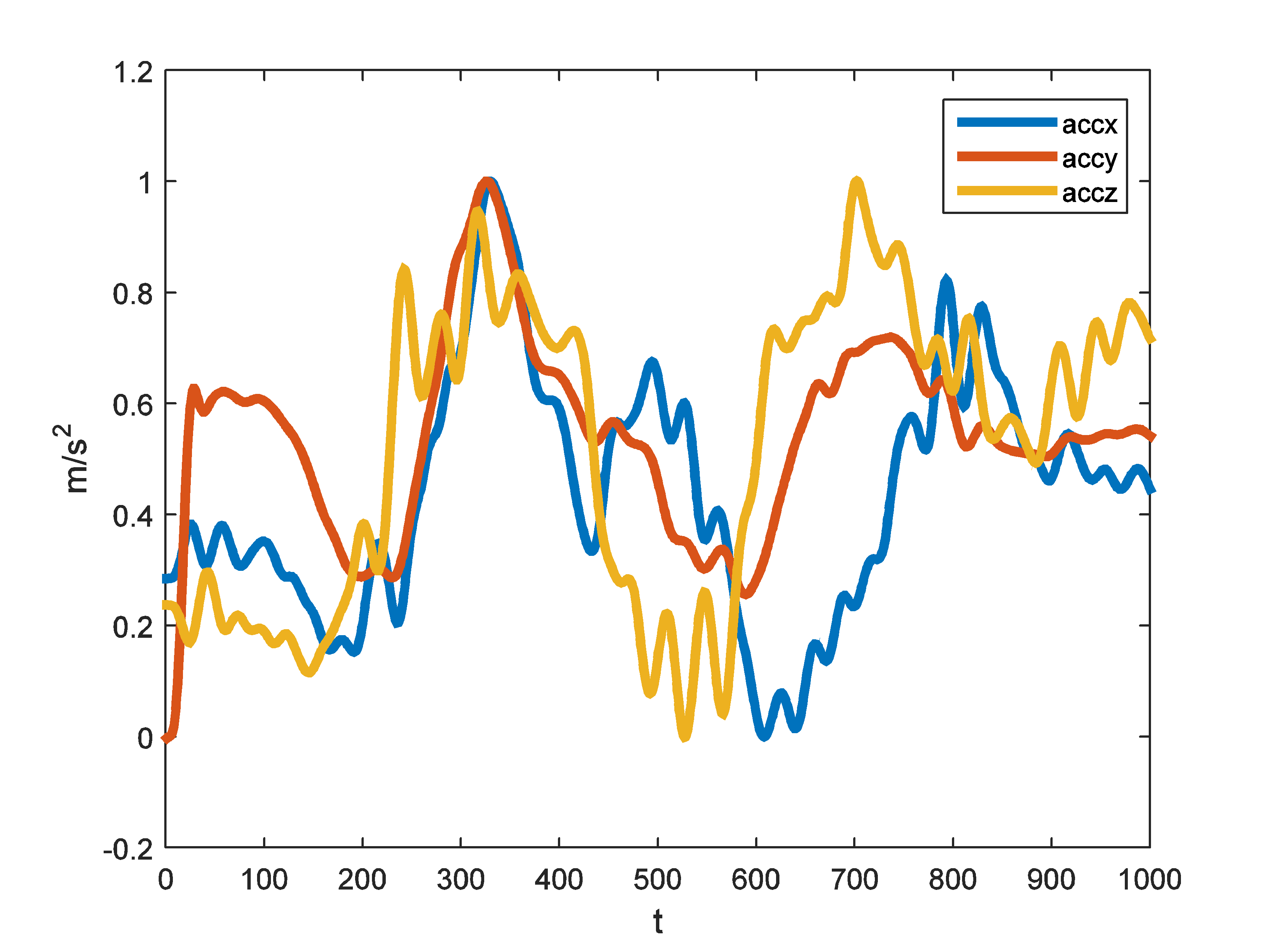}}
	\subfigure[Preprocessed Gyrometer Data]{
		\centering
		\includegraphics[width=0.47\linewidth]{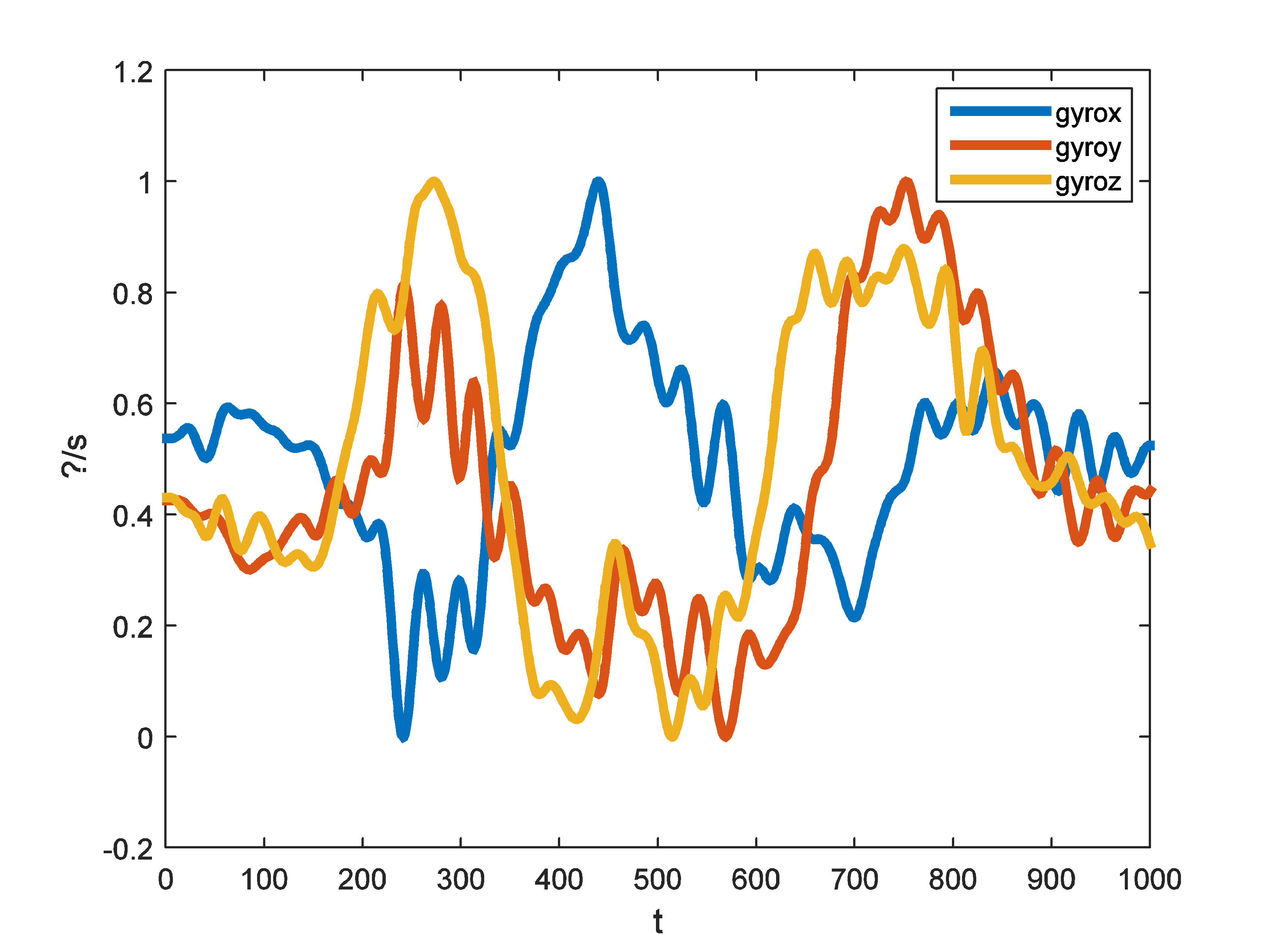}}
	\\
	\caption{The preprocessed accelerometer and gyrometer data.}
	\label{fig:preprocessed-data}
	\vspace*{4pt}
\end{figure*}

The gesture execution speed of different participants may vary considerably, which leads to different signal lengths due to a fixed sampling frequency (200HZ) of accelerometer and gyrometer in the mobile phone. For example, gestures completed relatively faster will have fewer sampling points. Also, the signal strength of gesture sequences may vary. To cope with signal strength and speed variations, we apply amplitude and sequence normalization to the original signal sequences. Specifically, we first normalize a signal $x_i^n\left(t\right)$ by
\begin{equation}
	x_i^n\left( t \right) = \frac{{{x_i}\left( t \right) - \min _{t = 1}^T{x_i}\left( t \right)}}{{\max _{t = 1}^T{x_i}\left( t \right) - \min _{t = 1}^T{x_i}\left( t \right)}},\;\;\;
	\forall i \in \left\{ {1,...,6} \right\}.
	\label{eq:normalize}
\end{equation}
Then, we use cubic spline interpolation to normalize the length of a sequence to a fixed size (we set this size as 1000 in our experiments). Fig.\;\ref{fig:preprocessed-data} shows the preprocessed accelerometer and gyrometer data, where the sequence has been filtered and normalized.

%---------------------------------------------------------
\subsection{Study of Fisher Criterion Parameters}\label{sec:parameter}

In our model, the hyperparameter $\theta$ impacts the Fisher criterion, $\alpha$ controls the update rate of mean $\mu$ in F-BLSTM, and $\delta$ adjusts the relationship of intra-class distance and inter-class distance between features. These parameters would affect the performance of gesture recognition. In order to configure an optimal parameter setting, we conduct parameter tuning experiments for the F-BLSTM model as follows.

\begin{figure*}[htp]
	\normalsize
	\begin{minipage}[t]{\linewidth}
		\centering
		\includegraphics[width=0.5\linewidth]{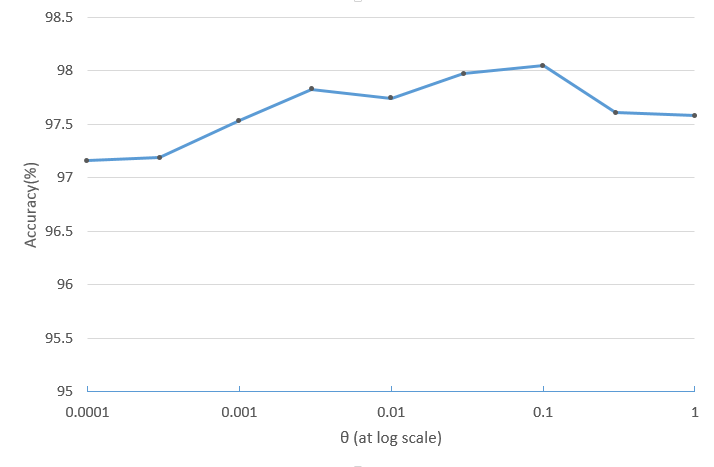}
		\caption{Influence of parameter $\theta$ on recognition accuracy.}
		\label{fig:parameter-theta}
	\end{minipage}
	\hfill \\
	\begin{minipage}[h]{\linewidth}
		\centering
		\includegraphics[width=0.5\linewidth]{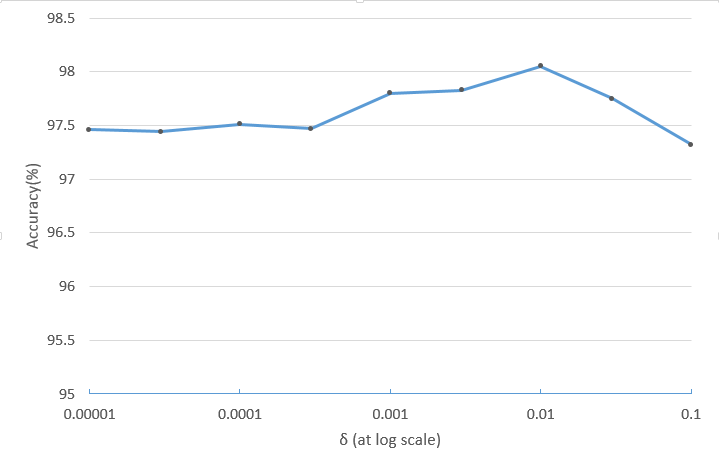}
		\caption{Influence of parameter $\delta$ on recognition accuracy.}
		\label{fig:parameter-delta}
	\end{minipage}
	\hfill
\end{figure*}

\begin{figure*}[h]
	\normalsize
	\begin{minipage}[t]{\linewidth}
		\centering
		\includegraphics[width=0.5\linewidth]{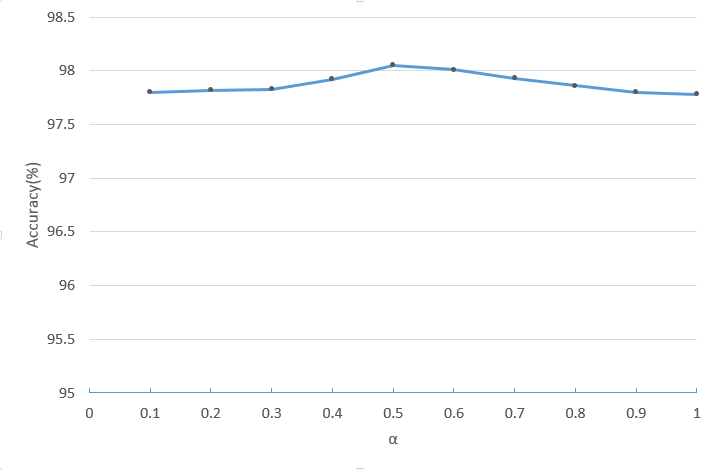}
		\caption{Influence of parameter $\alpha$ on recognition accuracy.}
		\label{fig:parameter-alpha}
	\end{minipage}
\end{figure*}

\begin{enumerate}[\textbf{Experiment} 1.]
	\item We fix $\alpha$ to 0.5, $\delta$ to 0.01 and vary $\theta$ from 0 to 1 to investigate the effect of parameter $\theta$. Fig.\;\ref{fig:parameter-theta} shows the classification accuracy on testing set. The result shows that the model trained with only softmax loss has poor performance, which certificates the necessity of introducing Fisher criterion.
	\item We fix $\alpha$ to 0.5, $\theta$ to 0.1 and vary $\delta$ from 1e-5 to 0.1 to verify that the inter-class distance’s joining promote the classification ability. As shown in Fig.\;\ref{fig:parameter-delta}, $\delta$  balances the relationship of intra-class distance and inter-class distance. We can set $\delta$ to an appropriate value to make the classification better according to different circumstances.
	\item We fix $\theta$ to 0.1, $\delta$ to 0.01 and vary $\alpha$ from 0 to 1 to train different models. The classification accuracy of these models on our gesture database are illustrated in Fig.\;\ref{fig:parameter-alpha}. We find that the classification performance of our model remains relatively stable across a wide range of $\alpha$, but a moderate value of $\alpha$ has a better performance.
\end{enumerate}

%---------------------------------------------------------
\begin{figure*}[h]
	\normalsize
	\centering
	\subfigure[BLSTM]{
		\centering
		\includegraphics[width=0.47\linewidth]{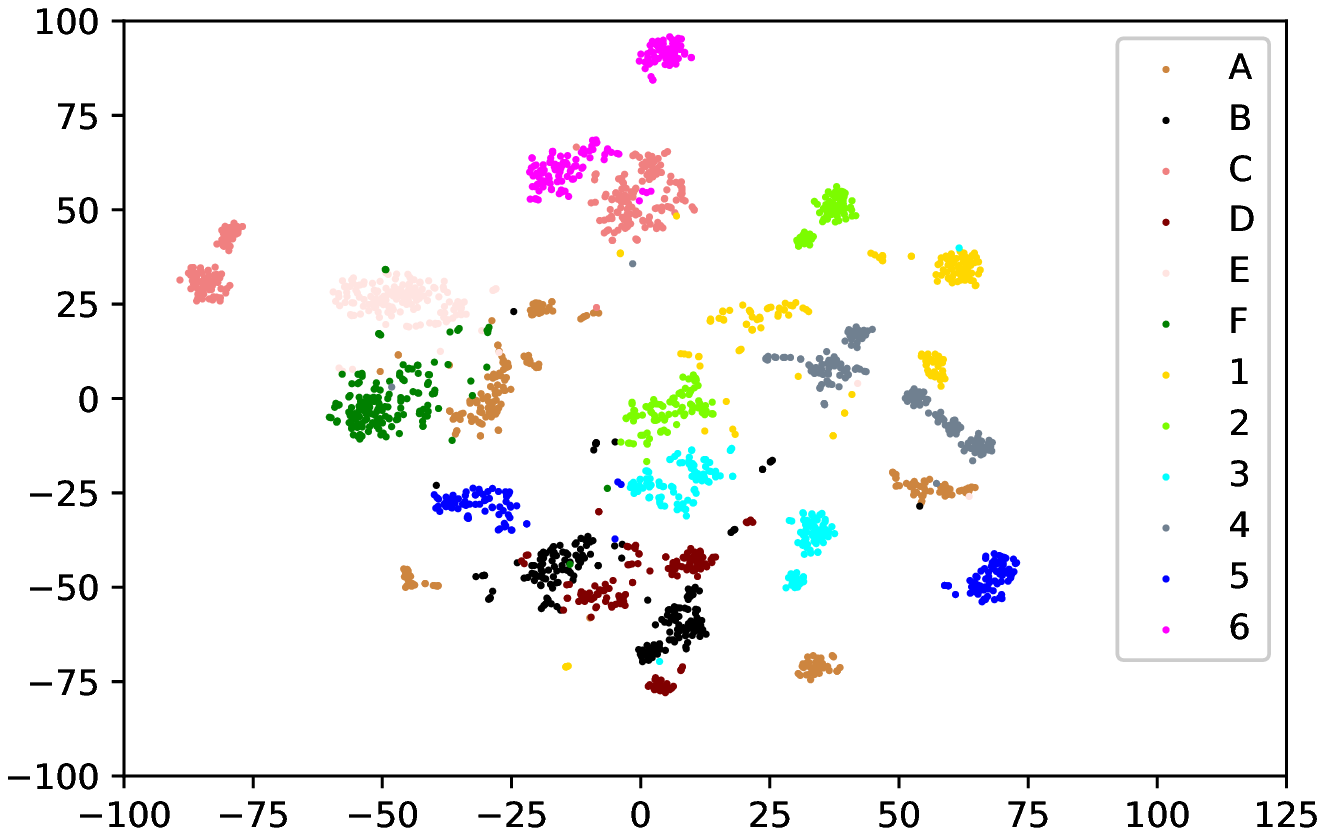}}
	\subfigure[BGRU]{
		\centering
		\includegraphics[width=0.47\linewidth]{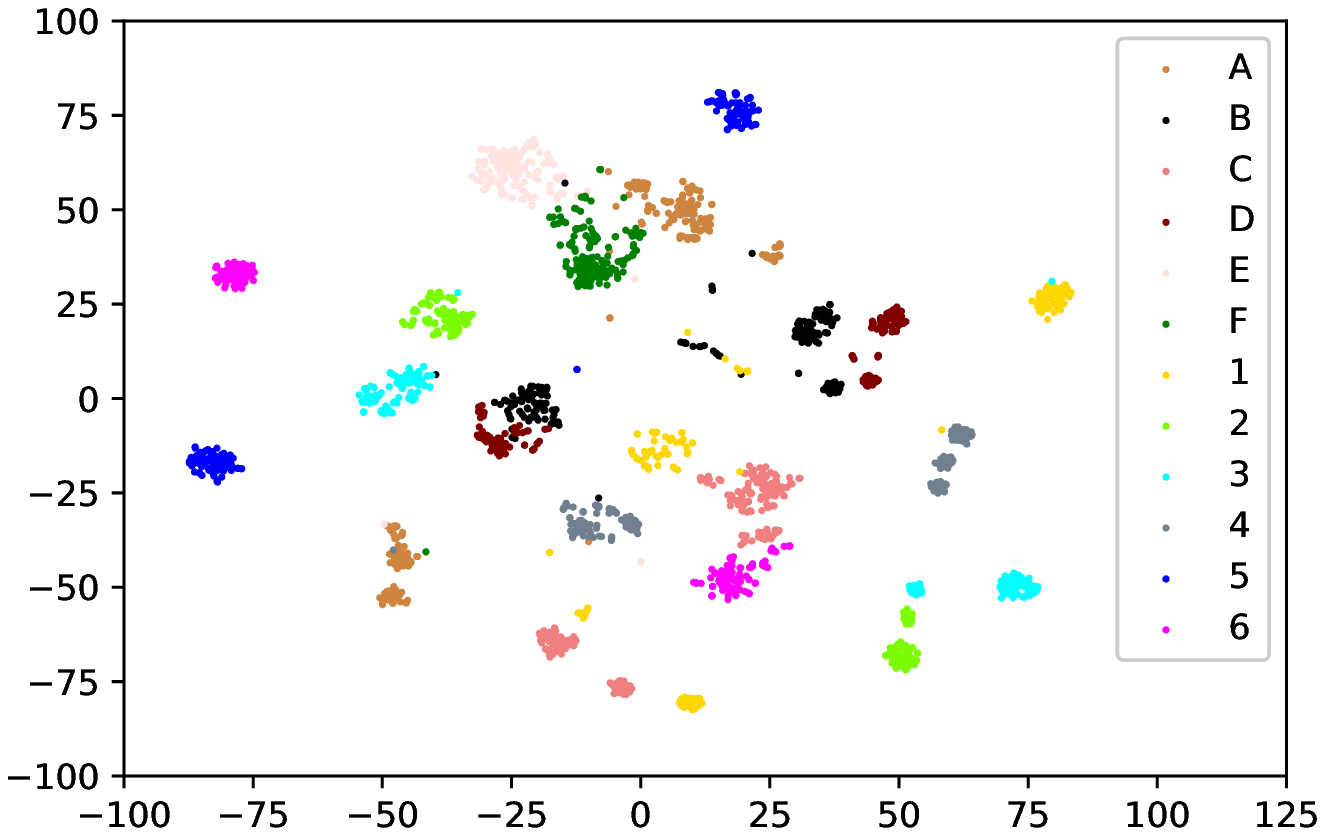}}
	\\
	\subfigure[F-BLSTM]{
		\centering
		\includegraphics[width=0.47\linewidth]{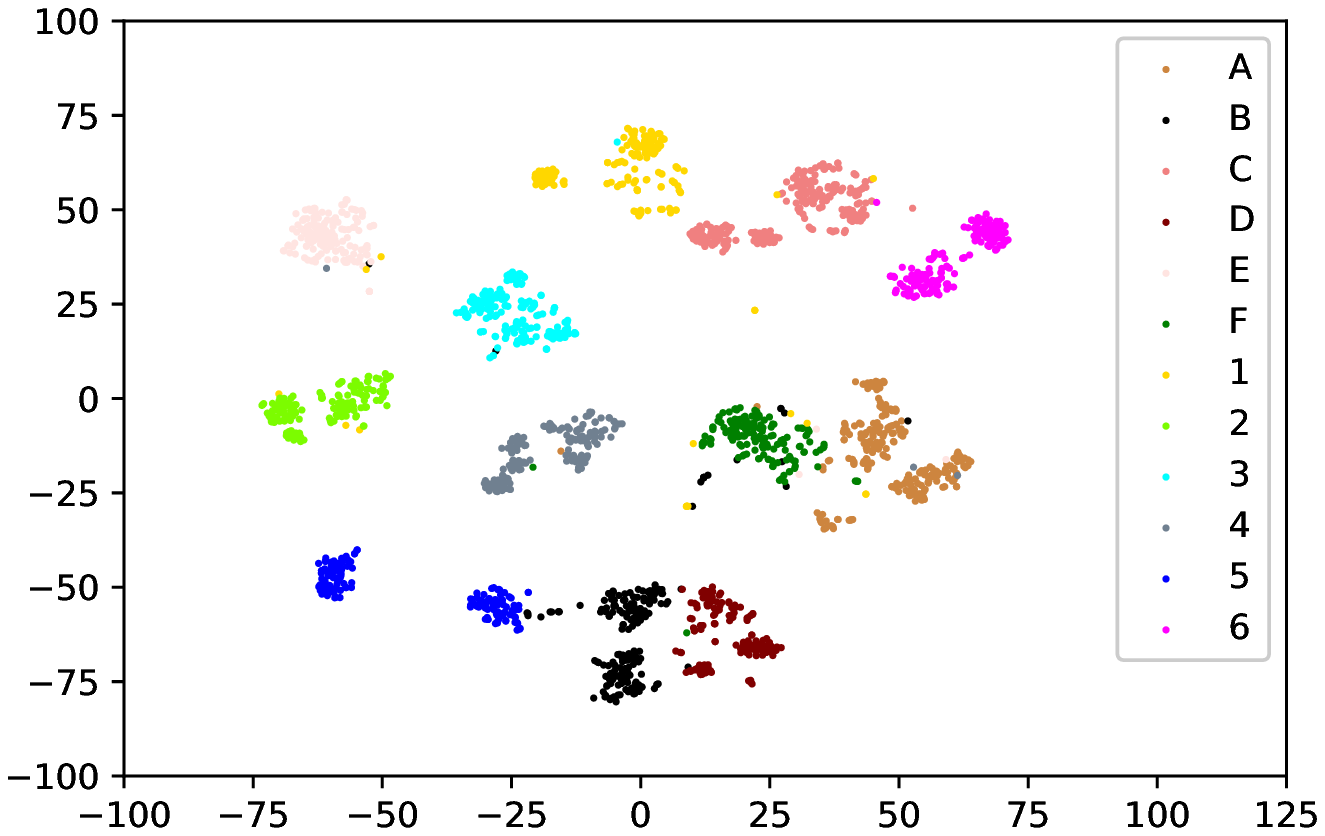}}
	\subfigure[F-BGRU]{
		\centering
		\includegraphics[width=0.47\linewidth]{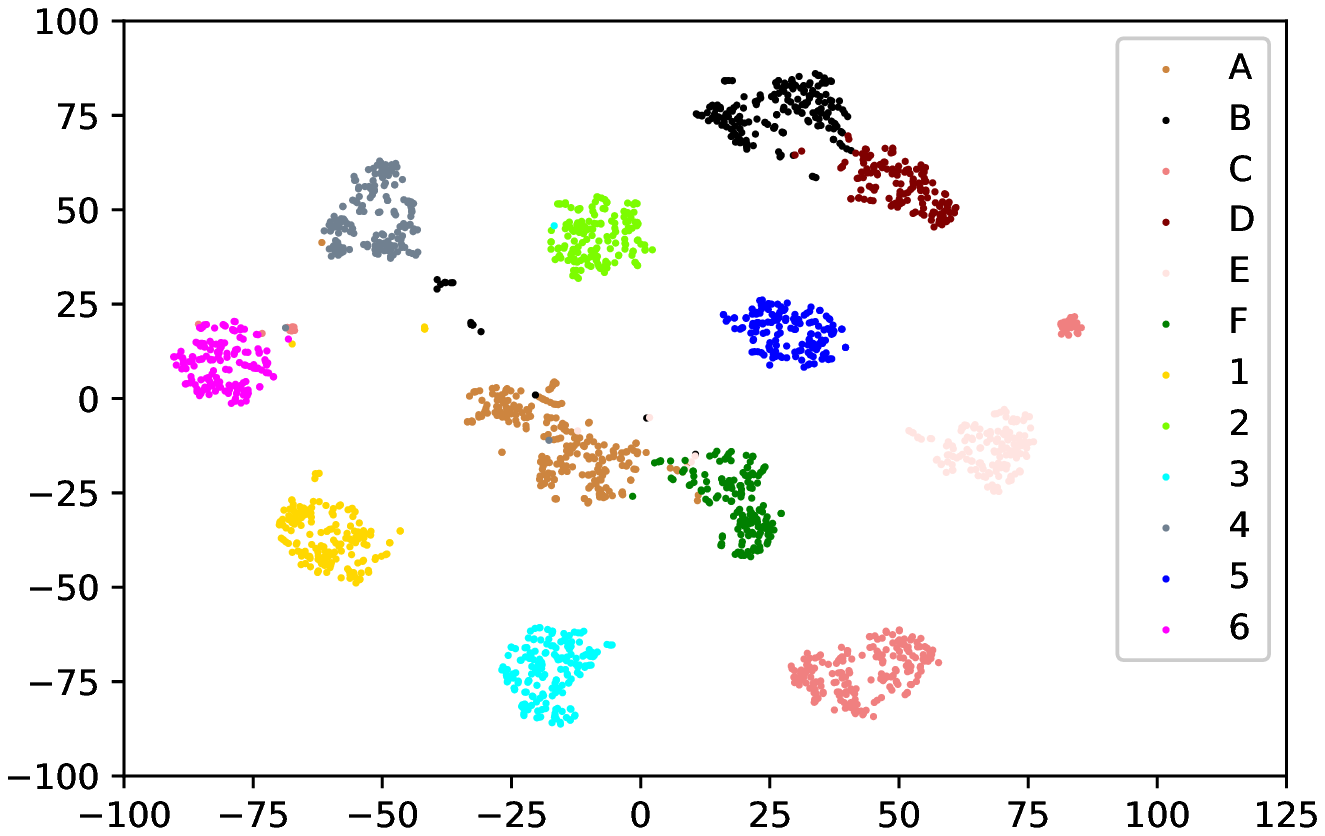}}
	\\
	\caption{Feature visualization of 12 classes of MGD database.}
	\label{fig:featurevisualization}
	\vspace*{4pt}
\end{figure*}
%\subsection{Discussion}\label{sec:discussion}
%---------------------------------------------------------
\subsection{Analysis of Model Effect}\label{sec:discussion}
As shown in the analysis mentioned above, the F-BLSTM and FBGRU achieves a better discriminant ability than the baseline BLSTM and BGRU. This section further discusses the better feature disctribution has been achieved. By the study in last section, we set $\theta$ to 0.1, $\delta$ to 0.01 and $\alpha$ to 0.5 in the F-BLSTM model, and set the parameters to 0.3, 0.01, 0.5 in the F-BGRU model, respectively.

Fig.~\ref{fig:featurevisualization} shows the feature visualizations on MGD database. In Fig.~\ref{fig:featurevisualization}(a) and Fig.~\ref{fig:featurevisualization}(b), the BLSTM and BGRU features of 12 classes are visualized by t-SNE~\cite{Maaten2009Science} with the parameters of initial dimension 100 and perplexity dimension 30, while the F-BLSTM and F-BGRU features are illustrated in Fig.~\ref{fig:featurevisualization}(c) and Fig.~\ref{fig:featurevisualization}(d), respectively. Clearly, the fisher discriminant learning features are more discriminative than the original baseline features, especially the F-BGRU feature in Fig.~\ref{fig:featurevisualization}(d) can be better discriminated than the BLSTM feature in Fig.~\ref{fig:featurevisualization}(a). As another verification, the quantitative evaluation is performed based on three databases in the next section.

%---------------------------------------------------------
\subsection{Comparison with State-of-the-Arts}\label{sec:comparison}
\begin{figure*}[htbp]
	\normalsize
	\centering
	\includegraphics[width=\linewidth]{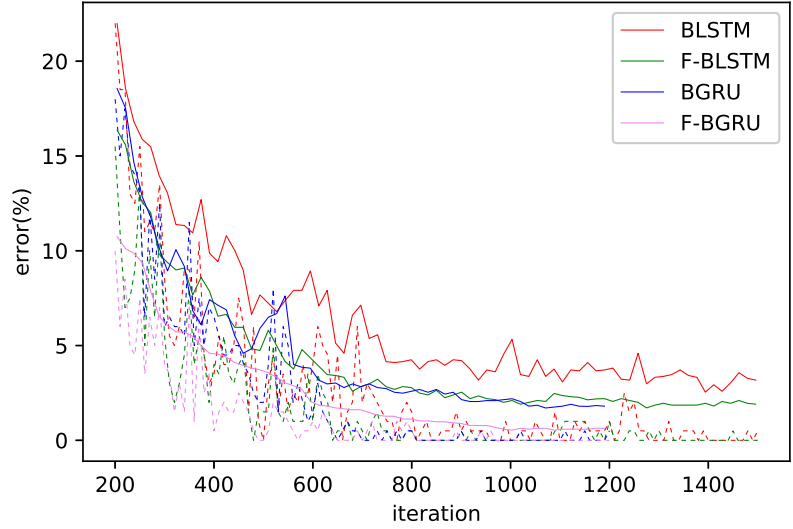}
	\caption{Training on \textbf{{MGD} database}. Dotted lines denote training errors, and solid lines denote testing errors. }
	\label{fig:curve-proposed}
	\vspace*{4pt}
\end{figure*}

\textbf{Experiment on {MGD} Database.}
For the proposed database, we select 3500 sequences for training our model and 2047 sequences for testing. After preprocessing, the length of each data sequence is set to 1000, thus each input sample (3-axis accelerometer and gyrometer signals) is a matrix of {$1000 \times 6$}. Here, we train the network using adaptive moment estimation, with the learning rate as 0.002 and the batch size as 200. For the F-BLSTM model, we set $\theta$ to 0.1, $\delta$ to 0.01 and $\alpha$ to 0.5. We complete the training of BLSTM and F-BLSTM models at 1.5K iterations. The parameters of F-BGRU model are set to 0.3, 0.01, 0.5 respectively. The training of BGRU and F-BGRU is completed at 1.2K iterations. We report the performance of different methods on the testing set based on the average over 5 runs. The class-wise classification accuracy comparison of different mothods is presented in Table.\;\ref{table:ourdatabase}. It is clear that by incorporating the Fisher criterion to the base models (BLSTM and BGRU), the recognition performance can be improved. Fig.\;\ref{fig:curve-proposed} also shows the behaviors of F-BLSTM and F-BGRU models. Dotted lines denote training errors, and solid lines denote testing errors for different methods. As can be observed from the figure, our proposed softmax function with Fisher criterion effectively speeds up the convergence of training and achieves a smaller error rate.

\begin{table*}[htp]
	\centering
	\caption{The average accuracy(\%) of BLSTM, BGRU and our proposed F-BLSTM, F-BGRU on \textbf{{MGD} database}.}
	\label{table:ourdatabase}
	\vspace*{6pt}
	\begin{tabular}{||c||c|c||c|c||}
		\hline
		\diagbox{\textbf{Gesture}}{\textbf{Method}} & BLSTM & F-BLSTM & BGRU & F-BGRU\\
		\hline
		A & 97.41 & 97.85 & 97.09 & 98.09 \\ %\hline
		B & 94.17 & 96.50 & 97.24 & 98.78 \\ %\hline
		C & 98.95 & 99.40 & 99.85 & 100.00 \\ %\hline
		D & 96.88 & 99.04 & 98.02 & 98.87 \\ %\hline
		E & 96.88 & 97.40 & 98.48 & 98.61 \\ %\hline
		F & 96.86 & 98.59 & 97.62 & 99.54 \\ %\hline
		1 & 93.80 & 95.33.82 & 96.62 & 98.53 \\ %\hline
		2 & 98.60 & 98.82 & 99.03 & 99.35 \\ %\hline
		3 & 96.69 & 97.56 & 98.29 & 99.42 \\ %\hline
		4 & 98.77 & 98.97 & 99.28 & 99.29 \\ %\hline
		5 & 96.55 & 98.16 & 99.77 & 100.00 \\ %\hline
		6 & 99.10 & 99.32 & 99.77 & 99.61 \\ \hline
		Overall & 97.05 & \textbf{98.04} & 98.38 & \textbf{99.15} \\ \hline
	\end{tabular}
\end{table*}

\begin{figure*}[htbp]
	\normalsize
	\centering
	\includegraphics[width=\linewidth]{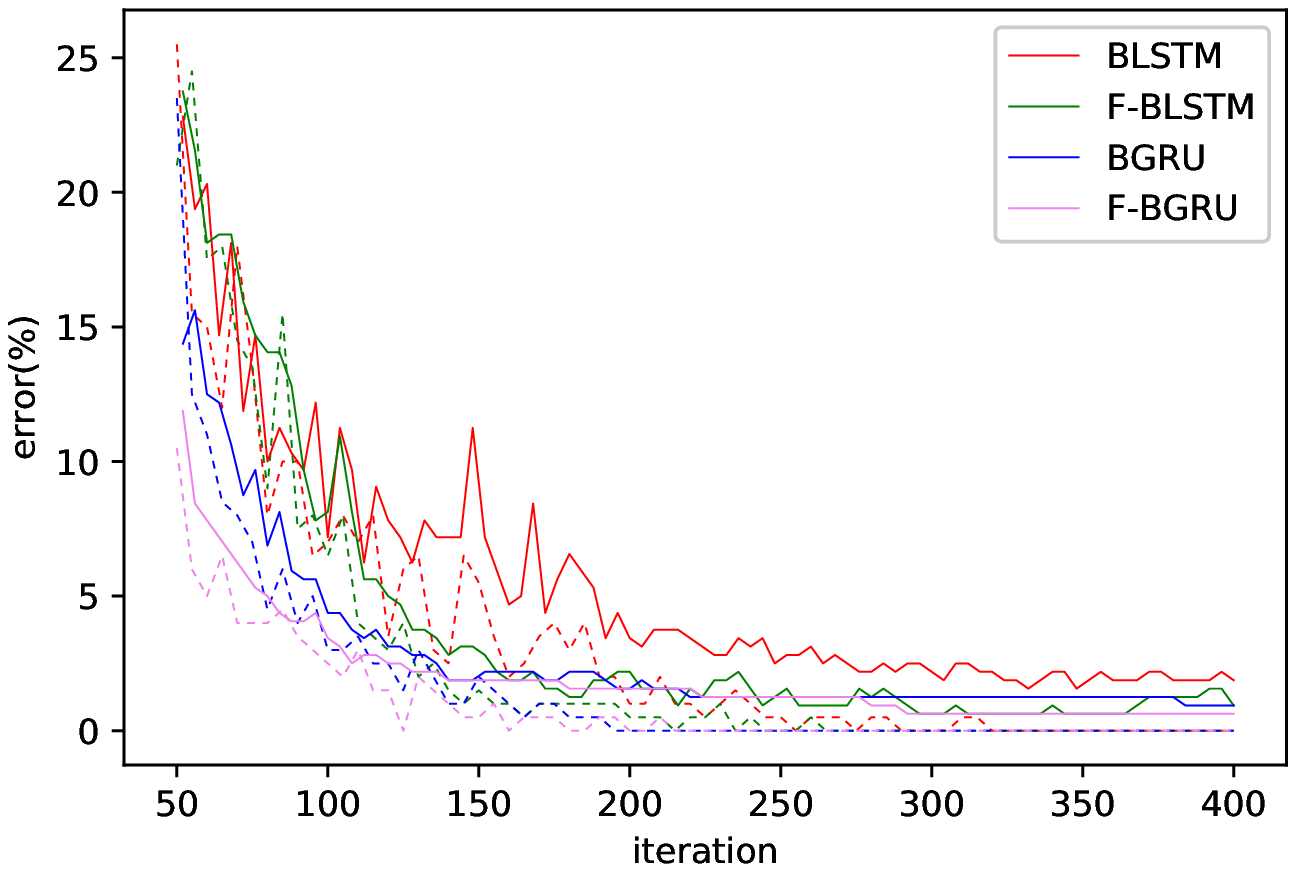}
	\caption{Training on \textbf{BUAA Mobile Gesture Database}. Dotted lines denote training errors, and solid lines denote testing errors. }
	\label{fig:curve-buaa}
	\vspace*{4pt}
\end{figure*}

\textbf{Experiment on BUAA Mobile Gesture Database\;\cite{xie2016ccbr}.}
%\subsubsection{Experiment on BUAA Mobile Gesture Database\;\cite{xie2016ccbr}}
This database has 1120 samples for gestures A, B, C, D, 1, 2, 3, 4. Each sample includes three-dimensional acceleration and angular velocity of the mobile phone. The training and testing sets are divided randomly into 70\% and 30\%, respectively. We conduct the experiments using the same setting for F-BLSTM and F-BGRU. We set $\theta$ to 0.1, $\delta$ to 0.03 and $\alpha$ to 0.5. Model training is finished at 400 iterations. Table.\;\ref{table:buaadatabase} shows that LSTMs with Fisher criterion still have better results than baselines on a smaller dataset. Similarly, the models converge faster and yield lower classification error rates with the Fisher criterion as shown in Fig.\;\ref{fig:curve-buaa}.

\begin{table*}
	\centering
	\caption{The average accuracy(\%) of BLSTM, BGRU and our proposed F-BLSTM, F-BGRU on \textbf{BUAA mobile gesture database}.}
	\label{table:buaadatabase}
	\vspace*{6pt}
	\begin{tabular}{||c||c|c||c|c||}
		\hline
		\diagbox{\textbf{Gesture}}{\textbf{Method}} & BLSTM & F-BLSTM & BGRU & F-BGRU\\
		\hline
		A & 100.00 & 99.17  & 98.34  & 99.58 \\ %\hline
		B & 97.29  & 98.92 & 97.84  & 98.37 \\ %\hline
		C & 100.00 & 100.00 & 100.00 & 100.00 \\ %\hline
		D & 99.26  & 97.42  & 96.77  & 99.35 \\ %\hline
		1 & 97.87  & 99.57  & 100.00 & 100.00 \\ %\hline
		2 & 100.00 & 100.00 & 100.00 & 100.00 \\ %\hline
		3 & 97.06  & 100.00 & 100.00 & 100.00 \\ %\hline
		4 & 95.83  & 97.50  & 97.08  & 97.08 \\ \hline
		Overall & 98.44 & \textbf{99.06} & 98.75 & \textbf{99.25} \\ \hline
	\end{tabular}
\end{table*}

\begin{figure*}[htbp]
	\normalsize
	\centering
	\includegraphics[width=\linewidth]{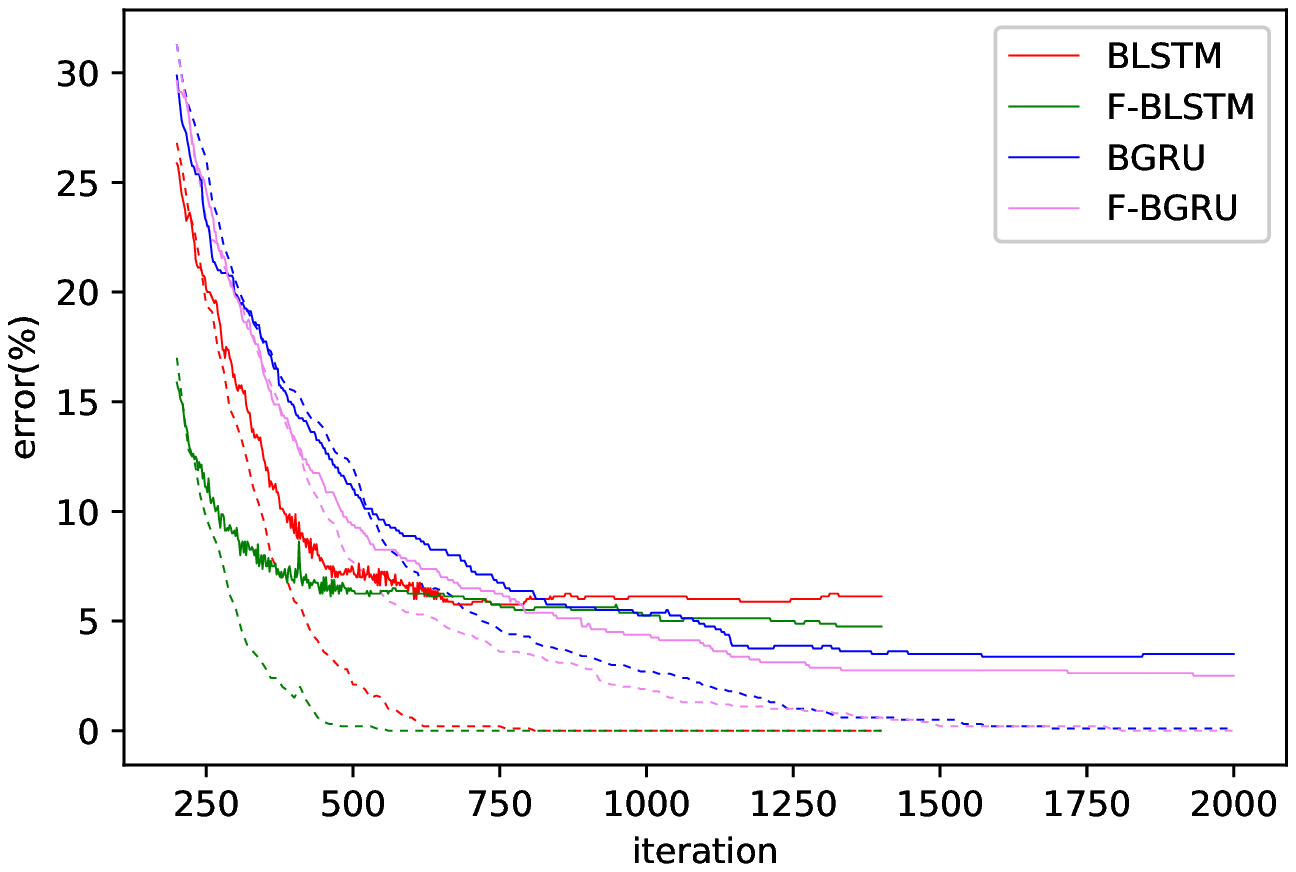}
	\caption{Training on \textbf{SmartWatch Gesture Database}. Dotted lines denote training errors, and solid lines denote testing errors. }
	\label{fig:curve-smart}
	\vspace*{4pt}
\end{figure*}

\textbf{Experiment on SmartWatch Gesture Database\;\cite{costante2014eusipco}.}
%\subsubsection{Experiment on SmartWatch Gesture Database\;\cite{costante2014eusipco}}
\begin{figure*}[htbp]
	\normalsize
	\centering
	\includegraphics[width=0.6\linewidth]{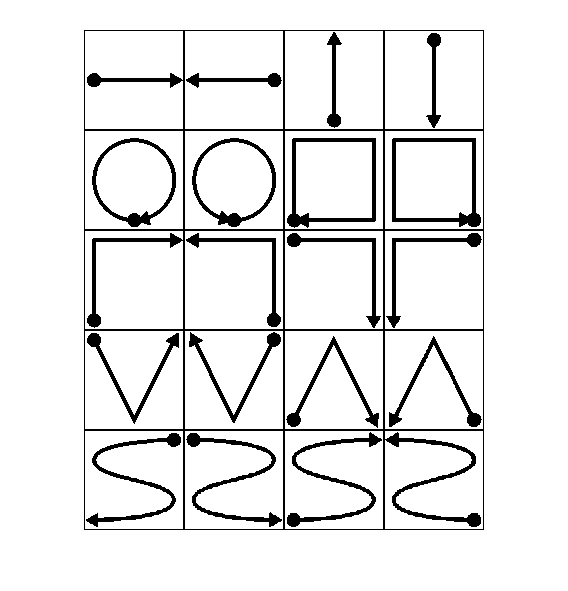}
	\\
	\caption{Examples of hand gestures in SmartWatch Database.}
	\label{fig:example-smartwatch}
	\vspace*{4pt}
\end{figure*}
The database has been used to evaluate gesture recognition algorithms for interacting with mobile applications using gestures. Eight different users performed twenty repetitions of twenty different gestures for a total of 3200 sequences. The gestures are depicted in Fig.\;\ref{fig:example-smartwatch}. Different from the 6-dimensional sequences of previous two databases, each sequence in this dataset only contains acceleration data from the 3-axis accelerometer of a first generation Sony SmartWatch. Furthermore, due to the lower sampling frequency, we set the length of each gesture sequence preprocessed to 50. We randomly select 2400 sequences as training set and the rest 800 sequences as testing set. The parameters of Fisher criterion follow the same setting in previous experiment. Adaptive moment estimation is used to train the network, and the initial learning rate $\lambda$ is set to 0.0001. The batch size is 1000. Training for BLSTM and F-BLSTM  is stopped at 1.4K iterations, and BGRU and F-BGRU at 2K iterations. {Fig.\;\ref{fig:curve-smart} shows the training error and testing error during the training process. Like Fig.\;\ref{fig:curve-proposed} and Fig.\;\ref{fig:curve-buaa}, dotted lines denote training errors, and solid lines denote testing errors. Table.\;\ref{table:smartwatch-database} lists the classification results for different gestures. Notice that our proposed models perform considerably better than the baselines across the 20 gestures.}
\begin{table*}
	\centering
	\caption{The average accuracy(\%) of BLSTM, BGRU and our proposed F-BLSTM, F-BGRU on \textbf{SmartWatch gesture database}.}
	\label{table:smartwatch-database}
	\vspace*{6pt}
	\begin{tabular}{||c||c|c||c|c||}
		\hline
		\diagbox{\textbf{Gesture}}{\textbf{Method}} & BLSTM & F-BLSTM & BGRU & F-BGRU\\
		\hline
		1  & 94.58  & 97.91 & 97.08  & 97.50 \\ %\hline
		2  & 95.00  & 97.22 & 95.56  & 95.56 \\ %\hline
		3  & 86.90  & 87.59 & 93.10  & 93.10 \\ %\hline
		4  & 95.91  & 97.27 & 97.27  & 97.73 \\ %\hline
		5  & 96.88  & 98.13 & 96.88  & 98.13 \\ %\hline
		6  & 93.33  & 94.07 & 96.30  & 100.00 \\ %\hline
		7  & 96.44  & 96.89 & 98.22  & 99.56 \\ %\hline
		8  & 97.62  & 98.57 & 100.00  & 100.00 \\ %\hline
		9  & 93.49  & 96.74 & 96.74  & 97.67 \\ %\hline
		10 & 94.84  & 98.06 & 100.00  & 100.00 \\ %\hline
		11 & 89.76  & 94.15 & 94.15  & 95.12 \\ %\hline
		12 & 92.89  & 92.44 & 96.00  & 97.33 \\ %\hline
		13 & 90.42  & 95.00 & 94.17  & 95.42 \\ %\hline
		14 & 94.88  & 96.30 & 96.30  & 97.21 \\ %\hline
		15 & 95.14  & 95.14 & 100.00  & 97.84 \\ %\hline
		16 & 92.20  & 89.27 & 93.17  & 93.17 \\ %\hline
		17 & 96.52  & 95.65 & 99.13  & 100.00 \\ %\hline
		18 & 96.22  & 97.30 & 96.76  & 95.68 \\ %\hline
		19 & 94.29  & 94.76 & 94.76  & 96.67 \\ %\hline
		20 & 97.21  & 98.60 & 100.00  & 100.00 \\ \hline
		Overall & 94.30 & \textbf{95.65} & 96.80 & \textbf{97.40} \\ \hline
	\end{tabular}
\end{table*}

Based on the experimental evaluations, we can observe that LSTM models with Fisher criterion consistently outperform the standard LSTMs on three different databases, validating the advantage of the proposed Fisher criterion. %The outstanding performance is not only reflected in BLSTM, but also in its variants. 
Furthermore, with even less data, the proposed Fisher criterion helps the LSTMs obtain a better result. This proves that it has a significant effect on a small database, showing the Fisher criterion has a widespread application scope.

%---------------------------------------------------------
\section{Conclusion}\label{sec:conclusion}
{In this paper, we have collected a large gesture database, namely MGD, for the application of mobile based gesture recognition. We proposed a novel Fisher criterion for F-BLSTM network to effectively classify the mobile hand gestures. Based on F-BLSTM, we also extended the F-BLSTM to a variant F-BGRU. By conducting numerous experiments, we proved that the mobile gesture recognition based on F-BLSTM networks which is supervised by the Fisher criterion has an outstanding performance. An appropriate values of Fisher criterion parameters contribute to the better results.}
%---------------------------------------------------------
\section*{Acknowledgement}
The work was supported in part by the Natural Science Foundation of China under Contract 61672079, 61473086, 61601466. The work of B. Zhang was supported in part by the Program for New Century Excellent Talents University within the Ministry of Education, China, and in part by the Beijing Municipal Science and Technology Commission under Grant Z161100001616005. Baochang Zhang is the correspondence.
%---------------------------------------------------------
\section*{References}
\bibliography{mybibfile}

\begin{thebibliography}{10}
\expandafter\ifx\csname url\endcsname\relax
  \def\url#1{\texttt{#1}}\fi
\expandafter\ifx\csname urlprefix\endcsname\relax\def\urlprefix{URL }\fi
\expandafter\ifx\csname href\endcsname\relax
  \def\href#1#2{#2} \def\path#1{#1}\fi

\bibitem{lane2010a}
N.~D. Lane, E.~Miluzzo, H.~Lu, D.~D. Peebles, T.~Choudhury, A.~T. Campbell, A
  survey of mobile phone sensing, IEEE Communications Magazine 48~(9) (2010)
  140--150.
\newblock \href {http://dx.doi.org/10.1109/MCOM.2010.5560598}
  {\path{doi:10.1109/MCOM.2010.5560598}}.

\bibitem{choi2005}
S.~C. J. Y. D. K. S.~K. Eunseok~Choi, Wonchul~Bang, Beatbox music phone:
  gesture-based interactive mobile phone using a tri-axis accelerometer, IEEE
  International Conference on Industrial Technology (2005) 97--102\href
  {http://dx.doi.org/10.1109/ICIT.2005.1600617}
  {\path{doi:10.1109/ICIT.2005.1600617}}.

\bibitem{mantyla2000hand}
V.~Mantyla, J.~Mantyjarvi, T.~Seppanen, E.~Tuulari, Hand gesture recognition of
  a mobile device user, International Conference on Multimedia and Expo 1
  (2000) 281--284.
\newblock \href {http://dx.doi.org/10.1109/ICME.2000.869596}
  {\path{doi:10.1109/ICME.2000.869596}}.

\bibitem{liu2009uwave}
J.~Liu, Z.~Wang, L.~Zhong, J.~Wickramasuriya, V.~Vasudevan, uwave:
  Accelerometer-based personalized gesture recognition and its applications,
  ieee international conference on pervasive computing and communications 5~(6)
  (2009) 1--9.
\newblock \href {http://dx.doi.org/10.1109/PERCOM.2009.4912759}
  {\path{doi:10.1109/PERCOM.2009.4912759}}.

\bibitem{catal2015on}
C.~Catal, S.~Tufekci, E.~Pirmit, G.~Kocabag, On the use of ensemble of
  classifiers for accelerometer-based activity recognition, Applied Soft
  Computing 37 (2015) 1018--1022.
\newblock \href {http://dx.doi.org/10.1016/j.asoc.2015.01.025}
  {\path{doi:10.1016/j.asoc.2015.01.025}}.

\bibitem{hochreiter1997long}
S.~Hochreiter, J.~Schmidhuber, Long short-term memory, Neural Computation 9~(8)
  (1997) 1735--1780.
\newblock \href {http://dx.doi.org/10.1162/neco.1997.9.8.1735}
  {\path{doi:10.1162/neco.1997.9.8.1735}}.

\bibitem{mikolov2011extensions}
T.~Mikolov, S.~Kombrink, L.~Burget, J.~Černocký, S.~Khudanpur, Extensions of
  recurrent neural network language model, in: 2011 IEEE International
  Conference on Acoustics, Speech and Signal Processing (ICASSP), 2011, pp.
  5528--5531.
\newblock \href {http://dx.doi.org/10.1109/ICASSP.2011.5947611}
  {\path{doi:10.1109/ICASSP.2011.5947611}}.

\bibitem{sundermeyer2012lstm}
M.~Sundermeyer, R.~Schluter, H.~Ney, Lstm neural networks for language
  modeling, Conference of the International speech Communication Association.

\bibitem{mesnil2013investigation}
G.~Mesnil, X.~He, L.~Deng, Y.~Bengio, Investigation of recurrent-neural-network
  architectures and learning methods for spoken language understanding,
  Conference of the International Speech Communication Association.

\bibitem{vinyals2015show}
O.~Vinyals, A.~Toshev, S.~Bengio, D.~Erhan, Show and tell: A neural image
  caption generator, CVPR (2015) 3156--3164\href
  {http://dx.doi.org/10.1109/CVPR.2015.7298935}
  {\path{doi:10.1109/CVPR.2015.7298935}}.

\bibitem{xu2015icml}
K.~Xu, J.~Ba, R.~Kiros, K.~Cho, A.~Courville, R.~Salakhutdinov, R.~Zemel,
  Y.~Bengio, Show, attend and tell: Neural image caption generation with visual
  attention, Computer Science (2015) 2048--2057.

\bibitem{ng2015cvpr}
J.~Y. Ng, M.~Hausknecht, S.~Vijayanarasimhan, O.~Vinyals, R.~Monga,
  G.~Toderici, Beyond short snippets: deep networks for video classification,
  CVPR (2015) 4694--4702\href {http://dx.doi.org/10.1109/CVPR.2015.7299101}
  {\path{doi:10.1109/CVPR.2015.7299101}}.

\bibitem{alahi2016social}
A.~Alahi, K.~Goel, V.~Ramanathan, A.~Robicquet, L.~Fei-Fei, S.~Savarese, Social
  lstm: Human trajectory prediction in crowded spaces, 2016 IEEE Conference on
  Computer Vision and Pattern Recognition (CVPR) (2016) 961--971\href
  {http://dx.doi.org/10.1109/CVPR.2016.110} {\path{doi:10.1109/CVPR.2016.110}}.

\bibitem{deng2015structure}
Z.~Deng, A.~Vahdat, H.~Hu, G.~Mori, Structure inference machines: Recurrent
  neural networks for analyzing relations in group activity recognition, 2016
  IEEE Conference on Computer Vision and Pattern Recognition (CVPR) (2016)
  4772--4781\href {http://dx.doi.org/10.1109/CVPR.2016.516}
  {\path{doi:10.1109/CVPR.2016.516}}.

\bibitem{ibrahim2015a}
M.~S. Ibrahim, S.~Muralidharan, Z.~Deng, A.~Vahdat, G.~Mori, A hierarchical
  deep temporal model for group activity recognition, 2016 IEEE Conference on
  Computer Vision and Pattern Recognition (CVPR) (2016) 1971--1980\href
  {http://dx.doi.org/10.1109/CVPR.2016.217} {\path{doi:10.1109/CVPR.2016.217}}.

\bibitem{du2015hierarchical}
Y.~Du, W.~Wang, L.~Wang, Hierarchical recurrent neural network for skeleton
  based action recognition, CVPR (2015) 1110--1118\href
  {http://dx.doi.org/10.1109/CVPR.2015.7298714}
  {\path{doi:10.1109/CVPR.2015.7298714}}.

\bibitem{veeriah2015differential}
V.~Veeriah, N.~Zhuang, G.~Qi, Differential recurrent neural networks for action
  recognition, ICCV (2015) 4041--4049\href
  {http://dx.doi.org/10.1109/ICCV.2015.460} {\path{doi:10.1109/ICCV.2015.460}}.

\bibitem{wang2014learning}
J.~Wang, Z.~Liu, Y.~Wu, J.~Yuan, Learning actionlet ensemble for 3d human
  action recognition, IEEE Transactions on Pattern Analysis and Machine
  Intelligence 36~(5) (2014) 914--927.
\newblock \href {http://dx.doi.org/10.1109/TPAMI.2013.198}
  {\path{doi:10.1109/TPAMI.2013.198}}.

\bibitem{liu2016eccv}
S.~A. Liu~J, X.~D, Spatio-temporal lstm with trust gates for 3d human action
  recognition, ECCV (2016) 816--833\href
  {http://dx.doi.org/10.1007/978-3-319-46487-9_50}
  {\path{doi:10.1007/978-3-319-46487-9_50}}.

\bibitem{shin2016dynamic}
S.~Shin, W.~Sung, Dynamic hand gesture recognition for wearable devices with
  low complexity recurrent neural networks, International Symposium on Circuits
  and Systems (2016) 2274--2277\href
  {http://dx.doi.org/10.1109/ISCAS.2016.7539037}
  {\path{doi:10.1109/ISCAS.2016.7539037}}.

\bibitem{lefebvre2013ANN}
G.~Lefebvre, S.~Berlemont, F.~Mamalet, C.~Garcia, Blstm-rnn based 3d gesture
  classification\href {http://dx.doi.org/10.1007/978-3-642-40728-4_48}
  {\path{doi:10.1007/978-3-642-40728-4_48}}.

\bibitem{rekimoto2001iswc}
J.~G. Rekimoto, Gesturepad, Unobtrusive wearable interaction devices, Fifth
  International Symposium on Wearable Computers (2001) 21--27\href
  {http://dx.doi.org/10.1109/ISWC.2001.962092}
  {\path{doi:10.1109/ISWC.2001.962092}}.

\bibitem{jang2003signal}
I.~J. Jang, W.~Park, Signal processing of the accelerometer for gesture
  awareness on handheld devices, Robot and Human Interactive Communication
  (2003) 139--144\href {http://dx.doi.org/10.1109/ROMAN.2003.1251823}
  {\path{doi:10.1109/ROMAN.2003.1251823}}.

\bibitem{kallio2003online}
S.~Kallio, J.~Kela, J.~Mantyjarvi, Online gesture recognition system for mobile
  interaction, Systems, Man and Cybernetics 3 (2003) 2070--2076.
\newblock \href {http://dx.doi.org/10.1109/ICSMC.2003.1244189}
  {\path{doi:10.1109/ICSMC.2003.1244189}}.

\bibitem{bulling2014a}
A.~Bulling, U.~Blanke, B.~Schiele, A tutorial on human activity recognition
  using body-worn inertial sensors, ACM Computing Surveys 46~(3) (2014) 33.
\newblock \href {http://dx.doi.org/10.1145/2499621}
  {\path{doi:10.1145/2499621}}.

\bibitem{ferscha2007gestural}
A.~Ferscha, S.~Resmerita, Gestural interaction in the pervasive computing
  landscape, Elektrotechnik Und Informationstechnik 124~(1) (2007) 17--25.
\newblock \href {http://dx.doi.org/10.1007/s00502-006-0413-4}
  {\path{doi:10.1007/s00502-006-0413-4}}.

\bibitem{Parsani2009A}
R.~Parsani, K.~Singh, A Single Accelerometer based Wireless Embedded System for
  Predefined Dynamic Gesture Recognition, Springer India, 2009.
\newblock \href {http://dx.doi.org/10.1007/978-81-8489-203-1_18}
  {\path{doi:10.1007/978-81-8489-203-1_18}}.

\bibitem{Roy2014Demo}
N.~Roy, H.~Wang, R.~{Roy Choudhury}, I am a smartphone and {I} can tell my
  user's walking direction, 2014, pp. 329--342.
\newblock \href {http://dx.doi.org/10.1145/2594368.2594392}
  {\path{doi:10.1145/2594368.2594392}}.

\bibitem{Park2014Poster}
J.~Park, K.~Kang, Intelligent classification of heartbeats for automated
  real-time ecg monitoring, Telemedicine journal and e-health : the official
  journal of the American Telemedicine Association 20~(12) (2014) 1069—1077.
\newblock \href {http://dx.doi.org/10.1089/tmj.2014.0033}
  {\path{doi:10.1089/tmj.2014.0033}}.

\bibitem{Nandakumar2015Contactless}
R.~Nandakumar, S.~Gollakota, N.~Watson, Contactless sleep apnea detection on
  smartphones, GetMobile: Mobile Comp. and Comm. 19~(3) (2015) 22--24.
\newblock \href {http://dx.doi.org/10.1145/2867070.2867078}
  {\path{doi:10.1145/2867070.2867078}}.

\bibitem{hoang2013adaptive}
T.~Hoang, T.~D. Nguyen, C.~Luong, S.~Do, D.~Choi, Adaptive cross-device gait
  recognition using a mobile accelerometer, Journal of Information Processing
  Systems 9~(2) (2013) 333--348.
\newblock \href {http://dx.doi.org/10.3745/JIPS.2013.9.2.333}
  {\path{doi:10.3745/JIPS.2013.9.2.333}}.

\bibitem{Agrawal2011Using}
S.~Agrawal, I.~Constandache, S.~Gaonkar, R.~R. Choudhury, K.~Caves,
  F.~Deruyter, Using mobile phones to write in air, in: International
  Conference on Mobile Systems, Applications, and Services, 2011, pp. 15--28.
\newblock \href {http://dx.doi.org/10.1145/1999995.1999998}
  {\path{doi:10.1145/1999995.1999998}}.

\bibitem{hofmann1998velocity}
F.~G. Hofmann, P.~Heyer, G.~Hommel, Velocity profile based recognition of
  dynamic gestures with discrete hidden markov models, Lecture Notes in
  Computer Science (1998) 81--95\href {http://dx.doi.org/10.1007/BFb0052991}
  {\path{doi:10.1007/BFb0052991}}.

\bibitem{markov}
S.~R. Eddy, Hidden markov models, Current opinion in structural biology 6~(3)
  (1996) 361--365.
\newblock \href {http://dx.doi.org/10.1016/S0959-440X(96)80056-X}
  {\path{doi:10.1016/S0959-440X(96)80056-X}}.

\bibitem{kela2006accelerometer-based}
J.~Kela, P.~Korpipaa, J.~Mantyjarvi, S.~Kallio, G.~Savino, L.~Jozzo, D.~Marca,
  Accelerometer-based gesture control for a design environment, Personal and
  Ubiquitous Computing 10~(5) (2006) 285--299.
\newblock \href {http://dx.doi.org/10.1007/s00779-005-0033-8}
  {\path{doi:10.1007/s00779-005-0033-8}}.

\bibitem{pylvanainen2005accelerometer}
T.~Pylvanainen, Accelerometer based gesture recognition using continuous hmms,
  iberian conference on pattern recognition and image analysis (2005)
  639--646\href {http://dx.doi.org/10.1007/11492429_77}
  {\path{doi:10.1007/11492429_77}}.

\bibitem{Zhang2009Hand}
X.~Zhang, X.~Chen, W.~H. Wang, J.~H. Yang, V.~Lantz, K.~Q. Wang, Hand gesture
  recognition and virtual game control based on 3d accelerometer and {EMG}
  sensors, 2009, pp. 401--406.
\newblock \href {http://dx.doi.org/10.1145/1502650.1502708}
  {\path{doi:10.1145/1502650.1502708}}.

\bibitem{akl2010accelerometer-based}
A.~Akl, S.~Valaee, Accelerometer-based gesture recognition via dynamic-time
  warping, affinity propagation, \& compressive sensing, International
  Conference on Acoustics, Speech, and Signal Processing (2010) 2270--2273\href
  {http://dx.doi.org/10.1109/ICASSP.2010.5495895}
  {\path{doi:10.1109/ICASSP.2010.5495895}}.

\bibitem{Mace2013Accelerometer}
D.~Mace, W.~Gao, A.~Coskun, Accelerometer-based hand gesture recognition using
  feature weighted na\"{\i}ve bayesian classifiers and dynamic time warping,
  in: Proceedings of the Companion Publication of the 2013 International
  Conference on Intelligent User Interfaces Companion, 2013, pp. 83--84.
\newblock \href {http://dx.doi.org/10.1145/2451176.2451211}
  {\path{doi:10.1145/2451176.2451211}}.

\bibitem{wu2009gesture}
J.~Wu, G.~Pan, D.~Zhang, G.~Qi, S.~Li, Gesture recognition with a 3-d
  accelerometer, Ubiquitous Intelligence and Computing (2009) 25--38\href
  {http://dx.doi.org/10.1007/978-3-642-02830-4_4}
  {\path{doi:10.1007/978-3-642-02830-4_4}}.

\bibitem{Hsu2009Integrating}
W.-H. Hsu, Y.-Y. Chiang, W.-Y. Lin, W.-C. Tai, J.-S. Wu, Integrating {LCS} and
  {SVM} for 3d handwriting recognition on handheld devices using
  accelerometers, in: Proceedings of the 3rd International Conference on
  Communications and Information Technology, 2009, pp. 195--197.

\bibitem{Marasovic2011Accelerometer}
V.~P. Tea~Marasovic, Accelerometer-based gesture classification using principal
  component analysis, in: SoftCOM 2011, 19th International Conference on
  Software, Telecommunications and Computer Networks, 2011, pp. 1 -- 5.

\bibitem{He2011Accelerometer}
Z.~He, Accelerometer based gesture recognition using fusion features and {SVM},
  JSW 6 (2011) 1042--1049.
\newblock \href {http://dx.doi.org/10.4304/jsw.6.6.1042-1049}
  {\path{doi:10.4304/jsw.6.6.1042-1049}}.

\bibitem{costante2014eusipco}
G.~Costante, L.~Porzi, O.~Lanz, P.~Valigi, E.~Ricci, Personalizing a
  smartwatch-based gesture interface with transfer learning, 2014 22nd European
  Signal Processing Conference (EUSIPCO) (2014) 2530--2534.

\bibitem{cho2014arxiv}
D.~B. K.~Cho, B. van~Merrienboer, Y.~Bengio, On the properties of neural
  machine translation: Encoder-decoder approaches, arXiv preprint 1409~(1259).

\bibitem{chung2014eprint}
C.~K. H. e.~a. Chung~J, Gulcehre~C, Empirical evaluation of gated recurrent
  neural networks on sequence modeling, Eprint arXiv.

\bibitem{xie2016ccbr}
C.~Xie, S.~Luan, H.~Wang, B.~Zhang, Gesture recognition benchmark based on
  mobile phone, CCBR\href {http://dx.doi.org/10.1007/978-3-319-46654-5_48}
  {\path{doi:10.1007/978-3-319-46654-5_48}}.

\bibitem{Maaten2009Science}
L.~V.~D. Maaten, E.~o. Postma, H.~J.~V.~D. Herik, Dimensionality reduction: A
  comparative review, IEEE Transactions on Pattern Analysis and Machine
  Intelligence 10.

\end{thebibliography}
%---------------------------------------------------------
\section*{Biography}
\textbf{Chunyu Xie}
	received the B.S. degree and is a master in automation from Beihang University. His current research interests include signal and image processing, pattern recognition and computer vision.\\ \\
\textbf{Ce Li}
	received the B.E. degree in Computer Science from Tianjin University, Tianjin, China, in 2008, the M.S. and Ph.D. degrees in Computer Science from the School of Electronic, Electrical and Communication Engineering at the University of Chinese Academy of Sciences, Beijing, China, in 2012 and 2015, respectively. She is currently a research assistant with China University of Mining \& Technology, Beijing, China. Her current interests include computer vision, video analysis and machine learning. She was supported by the Natural Science Foundation of China for Youth.\\ \\
\textbf{Baochang Zhang}
	received the B.S., M.S. and Ph.D. degrees in Computer Science from Harbin Institue of the Technology, Harbin, China, in 1999, 2001, and 2006, respectively. From 2006 to 2008, he was a research fellow with the Chinese University of Hong Kong, Hong Kong, and with Griffith University, Brisban, Australia. Currently, he is an associate professor with the Science and Technology on Aircraft Control Laboratory, School of Automation Science and Electrical Engineering, Beihang University, Beijing, China. He was supported by the Program for New Century Excellent Talents in University of Ministry of Education of China. His current research interests include pattern recognition, machine learning, face recognition, and wavelets.\\ \\
\textbf{Chen Chen}
	received the B.E. degree in automation from Beijing Forestry University, Beijing, China, in 2009, the M.S. degree in electrical engineering from Mississippi State University, Starkville, MS, USA, in 2012, and the Ph.D. degree from the University of Texas at Dallas, Richardson, TX, USA, in 2016. He is currently a Postdoctoral Fellow with the Center for Research in Computer Vision, University of Central Florida, Orlando, FL, USA. His current research interests include compressed sensing, signal and image processing, pattern recognition, and computer vision. He has published over 40 papers in refereed journals and conferences in the above areas.\\ \\
\textbf{Jungong Han}
	is currently a Senior Lecturer with the Department of Computer Science and Digital Technologies at Northumbria University, Newcastle, UK. Previously, he was a Senior Scientist (2012-2015) with Civolution Technology (a combining synergy of Philips Content Identification and Thomson STS), a Research Staff (2010-2012) with the Centre for Mathematics and Computer Science (CWI), and a Senior Researcher (2005-2010) with the Technical University of Eindhoven (TU/e) in Netherlands. Dr. Han’s research interests include Multimedia Content Identification, Multi-Sensor Data Fusion, Computer Vision and Multimedia Security. He is an Associate Editor of Elsevier Neurocomputing (IF 2.4) and an Editorial Board Member of Springer Multimedia Tools and Applications (IF 1.4).  He has been (lead) Guest Editor for five international journals, such as IEEE-T-SMCB, IEEE-T-NNLS. Dr. Han is the recipient of the UK Mobility Award Grant from the UK Royal Society in 2016. \\ \\
\end{document}